\documentclass[journal,transmag]{IEEEtran}

\ifCLASSINFOpdf
\usepackage{graphicx}  % enables \includegraphics

\usepackage{amsmath}
\usepackage{amssymb}

\begin{document}

\title{\textbf{Neuro-Symbolic Geometric Abstraction (NeuSOGA): From Observations to Symbolic Mathematical Representations}}

\author{
\IEEEauthorblockN{
Qingde Li\IEEEauthorrefmark{1},
Qingqi Hong\IEEEauthorrefmark{2},
Zihan Li\IEEEauthorrefmark{3},
Jie Tian\IEEEauthorrefmark{4},~\IEEEmembership{Fellow,~IEEE}}
\IEEEauthorblockA{\IEEEauthorrefmark{1}Computer Science, School of Digital and Physical Sciences, University of Hull, HU6 7RX, UK}
\IEEEauthorblockA{\IEEEauthorrefmark{2}The Institute of Artificial Intelligence, Xiamen University, Xiamen, China}
\IEEEauthorblockA{\IEEEauthorrefmark{3}Department of Bioengineering, University of Washington, Seattle, WA, USA}
\IEEEauthorblockA{\IEEEauthorrefmark{4} Chinese Academy of Sciences, Institute of Automation, Beijing, China}
}

% The paper headers
\markboth{Preprint. Under review.}%
{Shell \MakeLowercase{\textit{et al.}}: Bare Demo of IEEEtran.cls for IEEE Transactions on xxxx Journals}

\IEEEtitleabstractindextext{%

\begin{abstract} A fundamental challenge in artificial intelligence is the transformation of perceptual observations into explicit symbolic representations suitable for abstraction, reasoning, and knowledge formation. While modern AI systems achieve remarkable perceptual performance through large-scale statistical learning, the resulting knowledge is typically encoded within latent parameters that are difficult to interpret, manipulate, or analyse mathematically. Inspired by Neuro-Symbolic AI and theories of human abstraction, this paper investigates how symbolic mathematical representations can emerge from geometric observations. We propose NeuSOGA (Neuro-Symbolic Geometric Abstraction), a framework that implements the transformation  $\mathcal{O} \rightarrow \mathcal{T} \rightarrow \mathcal{G} \rightarrow \mathcal{S}$,  where observations ($\mathcal{O}$) are progressively transformed into topological abstractions ($\mathcal{T}$), geometric abstractions ($\mathcal{G}$), and ultimately symbolic mathematical representations ($\mathcal{S}$). The architecture combines topology-guided structural discovery using Euclidean Distance Transforms, foundation-model perception using Segment Anything, adaptive multi-scale geometric abstraction based on scale-space theory, and symbolic synthesis through Implicit Area Splines. Unlike conventional learning-based reconstruction approaches, structural knowledge is derived primarily through topological and geometric reasoning rather than task-specific geometric training. The resulting representation is an analytical implicit model possessing arbitrary-order smoothness ($C^n$ continuity), strict additivity, and closed-form evaluation. Unlike neural latent encodings, the generated Area Spline representation remains interpretable, editable, compositional, and mathematically explicit. Experiments on ModelNet40 point-cloud data, arbitrary-view projections, and segmented optical observations demonstrate that NeuSOGA consistently transforms diverse observations into compact symbolic representations while preserving essential topological and geometric structure across sensing modalities and viewing directions. The principal contribution of this work is the introduction of a neuro-symbolic mechanism for symbolic representation formation, establishing a computational bridge between perception and mathematical representation. By demonstrating the transformation from observation to symbol through topology-guided geometric abstraction, NeuSOGA provides an interpretable and explainable foundation for future systems capable of higher-level geometric reasoning, concept formation, and machine-assisted mathematical discovery. \end{abstract}

}

% make the title area
\maketitle

\begin{figure*}[t] \centering \includegraphics[width=\textwidth]{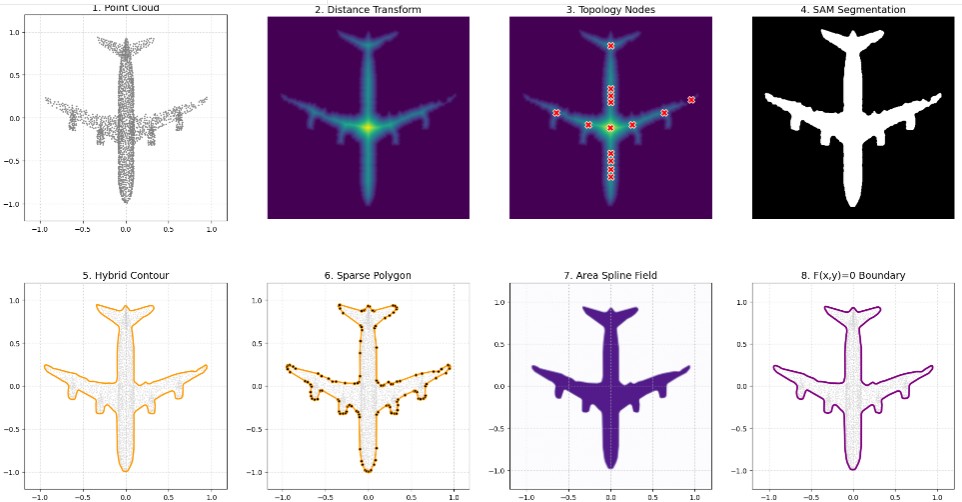} 
\caption{ Overview of the proposed NeuSOGA (Neuro-Symbolic Geometric Abstraction) pipeline: \mbox{$\mathcal{O}\rightarrow\mathcal{T}\rightarrow\mathcal{G}\rightarrow\mathcal{S}$}. Starting from a raw point-cloud observation, the framework extracts topology-aware structural cores using Euclidean Distance Transforms, employs these abstractions to guide foundation-model perception, performs adaptive multi-scale geometric abstraction, and finally constructs an explicit symbolic mathematical representation through Implicit Area Spline synthesis. Unlike conventional learning-based reconstruction methods that encode geometry within latent neural parameters, the proposed framework produces an analytical representation $F(x,y)=0$ that is interpretable, editable, additive, and mathematically explicit. }
\label{fig:overview} \end{figure*}

\IEEEdisplaynontitleabstractindextext

\IEEEpeerreviewmaketitle

\section{Introduction}
Recent advances in deep learning have produced powerful models capable of perception, pattern recognition, and geometric inference across a wide range of domains, including computer vision, shape analysis, and computer-aided design reconstruction \cite{bommasani2021opportunities,park2019deepsdf,wu2021deepcad,khan2024cad,liu2024point2cad}. These systems typically derive their capabilities from large-scale statistical learning, where neural networks are optimized using vast collections of training examples. While this paradigm has achieved remarkable practical success, the resulting representations are generally encoded within latent neural parameters whose structure is difficult to interpret, manipulate, or analyse directly. Furthermore, generalization remains challenging when observations differ substantially from the geometries, structures, or topologies encountered during training \cite{lake2017building,marcus2020next}. The computational and energy demands associated with large-scale neural systems have also motivated growing interest in alternative approaches that combine learning with more explicit forms of representation and reasoning \cite{strubell2019energy,garcez2022neurosymbolic}. Human mathematical cognition appears to operate in a fundamentally different manner. Rather than relying solely on large numbers of examples, humans can often infer general geometric principles from relatively few observations and subsequently construct abstract symbolic representations that extend beyond immediate sensory experience \cite{lake2017building,tenenbaum2011grow}. Developmental cognitive research suggests that humans possess innate forms of ``core knowledge'' concerning objects, space, geometry, and topological relationships, providing a foundation for higher-level abstraction and reasoning \cite{spelke2007core,geary2000evolution}. Lake et al. argue that human intelligence is characterized not only by pattern recognition but also by its ability to construct causal, compositional, and abstract representations supporting systematic generalization \cite{lake2017building}. Such capabilities allow mathematicians and scientists to derive idealized geometric structures and analytical models from incomplete and noisy observations of the physical world. This observation motivates a broader view of intelligence as a hierarchical process of abstraction. At the perceptual level, observations are acquired from an uncertain and noisy environment. At the structural level, invariant geometric and topological relationships are extracted from these observations. At the symbolic level, the resulting structures are represented analytically in forms that support interpretation, manipulation, and subsequent reasoning \cite{newell1976computer}. From this perspective, mathematical abstraction may be viewed as a process of transforming observations into explicit symbolic representations through successive layers of structural organization. The growing interest in neuro-symbolic artificial intelligence reflects the recognition that neither purely neural nor purely symbolic approaches are independently sufficient for supporting this form of abstraction \cite{garcez2002neural,bader2005dimensions, garcez2015reasoning,besold2017survey, garcez2022neurosymbolic}. Neural systems provide robustness to noise, ambiguity, and incomplete information, while symbolic systems offer explicit representations, interpretability, compositionality, and analytical structure \cite{newell1976computer}. A neuro-symbolic architecture therefore provides a promising framework for investigating how symbolic representations can emerge from perceptual observations. The Scan-to-CAD problem provides a particularly useful testbed for exploring this question. The objective is to transform unstructured geometric observations into compact, editable, and mathematically meaningful representations. Recent approaches such as DeepCAD, CAD-SIGNet, Point2CAD, and related CAD reconstruction systems have demonstrated impressive performance through the learning of geometric priors from large collections of synthetic CAD models \cite{wu2021deepcad,khan2024cad,liu2024point2cad, mallis2023sharp}. However, these methods remain fundamentally dependent upon learned statistical representations and training data distributions. Consequently, generalization to previously unseen topologies, sparse observations, and structurally novel geometries remains an active challenge. We argue that a key limitation of existing approaches is the absence of an explicit abstraction layer that separates topological structure, geometric form, and symbolic representation. A point cloud contains not only geometric measurements but also latent structural information that reflects the organization of the observed object. Rather than learning this structure exclusively through statistical inference, we propose to extract it directly through deterministic geometric operators. The Medial Axis Transform (MAT) introduced by Blum provides a compact description of shape topology through skeletal representations \cite{blum1967transformation}, while Euclidean Distance Transforms (EDT) provide an efficient mechanism for computing intrinsic spatial relationships within geometric domains \cite{maurer2003linear}. Together, these methods reveal topological abstractions that are largely independent of specific geometric parameterizations. Building upon this foundation, we introduce \textbf{NeuSOGA} (\textbf{Neu}ro-\textbf{S}ymbolic \textbf{G}eometric \textbf{A}bstraction), a computational framework that systematically transforms perceptual observations into explicit, editable symbolic mathematical representations via a structured hierarchy:
\begin{equation}
\mathcal{O} \xrightarrow{\text{Topology}} \mathcal{T} \xrightarrow{\text{Geometry}} \mathcal{G} \xrightarrow{\text{Synthesis}} \mathcal{S},
\end{equation}
where $\mathcal{O}$ denotes observational inputs (e.g., point clouds, binary silhouettes), $\mathcal{T}$ represents topological skeletal abstractions, $\mathcal{G}$ indicates sparse geometric control polygons, and $\mathcal{S}$ corresponds to analytical implicit spline fields.

Within NeuSOGA, foundation vision models (SAM/SAM2) \cite{kirillov2023segment,ravi2024sam2} operate as perceptual front ends, while structural understanding emerges deterministically through Euclidean Distance Transforms (EDT) and multi-scale scale-space analysis \cite{lindeberg1994scale,douglas1973algorithms}. The extracted geometric primitives are subsequently synthesized into closed-form Implicit Area Splines \cite{li20092d}. Unlike conventional parametric representations (e.g., NURBS \cite{piegl1997nurbs}) that encode boundary trajectories, Area Splines represent 2D regions directly through analytical implicit fields, providing arbitrary-order ($C^n$) smoothness, closed-form evaluation, and exact algebraic additivity. The principal contributions of this work are summarized as follows:
\begin{enumerate}
    \item \textbf{A Neuro-Symbolic Abstraction Hierarchy ($\mathcal{O}\rightarrow\mathcal{T}\rightarrow\mathcal{G}\rightarrow\mathcal{S}$):} We establish a training-free computational framework bridging unstructured perception and explicit mathematical modeling through topology-guided geometric abstraction.
    \item \textbf{Topology-Driven Perceptual Grounding:} We develop a deterministic prompting strategy that extracts maximal inscribed structural cores via Euclidean Distance Transforms to guide foundation segmentation models without task-specific tuning.
    \item \textbf{Adaptive Multi-Scale Geometric Simplification:} We formulate an error-bounded coarse-to-fine contour fusion mechanism based on scale-space theory, selectively retaining high-curvature engineering details while compressing boundaries into sparse polygons.
    \item \textbf{Analytical Synthesis via Implicit Area Splines:} We introduce Area Splines as an explicit symbolic representation layer, demonstrating that closed-form algebraic fields enable direct geometric editing, exact additivity, and continuous $C^n$ modeling.
    \item \textbf{Cross-Modality and Viewpoint Invariance:} We demonstrate through comprehensive evaluations on ModelNet40 point clouds (canonical and arbitrary views) and COCO optical masks that NeuSOGA achieves consistent symbolic abstraction across sensing modalities.
\end{enumerate}

Figure~\ref{fig:overview} illustrates the end-to-end architecture of NeuSOGA, detailing each progressive stage from initial observation to the final zero-level set $F(x,y)=0$.

\section{Related Work}

\subsection{Foundation Models, Deep Learning, and the Limits of Statistical Learning} Recent advances in deep learning and foundation models have transformed computer vision, natural language processing, and geometric understanding through large-scale data-driven learning \cite{bommasani2021opportunities}. Vision foundation models such as Segment Anything (SAM) \cite{kirillov2023segment} and SAM2 \cite{ravi2024sam2} demonstrate remarkable zero-shot generalization capabilities and provide highly effective mechanisms for extracting semantic information from visual data. Despite these successes, concerns remain regarding the ability of purely statistical learning systems to perform robust abstraction, reasoning, and knowledge discovery. Lake et al. argue that human intelligence depends upon compositionality, causal reasoning, and model-based abstraction, capabilities that are difficult to achieve through pattern recognition alone \cite{lake2017building}. Similarly, Marcus identifies systematic generalization, explainability, and reasoning as fundamental challenges for purely connectionist systems \cite{marcus2020next}. Furthermore, scaling modern deep learning systems increasingly requires substantial computational resources and energy consumption \cite{strubell2019energy}. These limitations have motivated significant interest in hybrid paradigms capable of combining the strengths of learning and reasoning. \subsection{Neuro-Symbolic AI and Machine Abstraction} The integration of neural learning and symbolic reasoning has been an active research area for more than two decades. Early neural-symbolic systems sought to combine the adaptability of neural networks with the formal expressiveness and inferential capabilities of symbolic logic \cite{garcez2002neural}. Bader and Hitzler classified neural-symbolic architectures according to their mechanisms for knowledge representation, learning, and reasoning, providing one of the earliest systematic frameworks for integrating symbolic and connectionist computation \cite{bader2005dimensions}. More recent work has expanded this vision into the broader field of Neuro-Symbolic AI. Garcez et al. argued that future AI systems require the principled integration of learning, reasoning, and knowledge representation \cite{garcez2015reasoning}. Besold et al. further identified neural-symbolic learning as a mechanism for bridging low-level representation learning and high-level cognitive reasoning \cite{besold2017survey}. This perspective has culminated in the concept of the ``third wave'' of AI, in which neural computation and symbolic reasoning are treated as complementary components of a unified intelligent system \cite{garcez2022neurosymbolic}. The motivation for neuro-symbolic AI is closely related to findings from cognitive science. Spelke and Kinzler's Core Knowledge theory suggests that humans possess structured priors concerning objects, geometry, space, and topology that support rapid abstraction from limited experience \cite{spelke2007core}. Similarly, Lake et al. propose that human intelligence emerges through the construction of compositional and reusable abstract models capable of supporting systematic generalization \cite{lake2017building}. These observations suggest that mathematical cognition relies on more than statistical learning alone; it requires mechanisms for extracting invariant structures and transforming them into symbolic representations. Our work adopts this perspective and investigates a topology-driven neuro-symbolic architecture in which perception, abstraction, and symbolic synthesis are explicitly separated. Unlike conventional learning-based approaches, abstraction is performed through deterministic geometric analysis rather than data-driven optimization. \subsection{Learning-Based Scan-to-CAD Reconstruction} The reconstruction of editable CAD models from point clouds remains a central problem in reverse engineering and geometric modeling. Benchmark initiatives such as the SHARP Challenge highlight the difficulty of recovering CAD design intent, construction history, and parametric information from real-world scans \cite{mallis2023sharp}. Early work such as CSGNet formulated reconstruction as the inference of constructive solid geometry programs from observed shapes \cite{sharma2018csgnet}. Subsequently, DeepCAD represented CAD reconstruction as a sequence-generation problem, learning parametric modeling operations from large CAD repositories \cite{wu2021deepcad}. Recent methods have introduced increasingly sophisticated architectures. CAD-SIGNet employs sketch-guided attention mechanisms to infer feature histories from point clouds \cite{khan2024cad}, while Point2CAD combines neural predictions with geometric fitting procedures to improve reconstruction accuracy \cite{liu2024point2cad}. KP-RED further utilizes semantically meaningful keypoints to establish geometric correspondence and improve shape recovery \cite{zhang2024kpred}. Although these approaches differ substantially in their architectures, they share a common assumption: geometric understanding is acquired through statistical learning from large collections of synthetic examples. Consequently, their ability to generalize remains dependent upon the diversity and completeness of the training distribution. In contrast, the proposed framework seeks to derive geometric structure directly from topological and geometric principles rather than from learned statistical correlations. \subsection{Topological Shape Abstraction} A fundamental requirement for mathematical abstraction is the discovery of invariant structure. Within geometric modeling, topology provides such invariants by describing structural relationships that remain unchanged under continuous deformation. Blum's Medial Axis Transform (MAT) represents one of the most influential approaches to topological shape analysis \cite{blum1967transformation}. By reducing a shape to its skeletal structure and associated radius function, MAT provides a compact representation of object organization while preserving essential topological information. Euclidean Distance Transforms (EDTs) complement this representation by providing efficient computation of intrinsic geometric relationships across arbitrary dimensions \cite{maurer2003linear}. Unlike learned latent representations, which often encode topology implicitly within network parameters, MAT and EDT provide explicit and interpretable descriptions of geometric structure. In the proposed framework, these methods serve as mechanisms for invariant extraction, transforming raw observations into symbolic topological descriptions that can subsequently support analytical reasoning. \subsection{Explicit and Implicit Geometric Representations} The representation of geometry has long been a central concern within CAD and solid modeling. Classical geometric systems rely primarily on explicit representations such as boundary representations (B-Reps), constructive solid geometry (CSG), and spline-based surface models \cite{requicha1980representations}. Among these, NURBS have become the dominant industrial standard due to their geometric flexibility, precision, and compatibility with manufacturing processes \cite{piegl1997nurbs}. However, explicit representations often require sophisticated numerical procedures to maintain topological consistency during Boolean operations, intersection calculations, and geometric editing \cite{requicha1980representations}. To overcome these limitations, implicit representations describe shapes as level sets of continuous scalar fields, enabling topological operations to be expressed naturally. Recent neural implicit approaches such as DeepSDF have demonstrated the representational power of learned continuous distance fields \cite{park2019deepsdf}. Nevertheless, geometric knowledge in such systems is encoded within network weights, limiting interpretability and direct integration with conventional design workflows. In contrast, Piecewise Algebraic Splines provide a fully analytical implicit representation in which geometry remains explicitly controllable through editable parameters while preserving the topological robustness of implicit modeling \cite{li20092d}. \subsection{Scale-Space Theory and Multi-Scale Geometric Reasoning} A central problem in geometric abstraction is the separation of meaningful structural information from noise and small-scale variations. Scale-space theory provides a rigorous mathematical framework for addressing this challenge by representing signals and geometric structures across a continuum of observation scales \cite{witkin1983scale,koenderink1984structure}. The fundamental principle underlying scale-space analysis is that meaningful structures should persist as the observation scale increases, while insignificant details gradually disappear. The mathematical foundations of scale-space theory were formalized by Lindeberg, who established a comprehensive framework for multi-scale representation, feature extraction, and scale invariance \cite{lindeberg1994scale,lindeberg1998scale}. One of the key insights of this framework is that no single observation scale is sufficient to capture all meaningful information. Instead, stable geometric structures emerge through comparisons across multiple scales. Lindeberg further demonstrated that automatic scale selection can identify characteristic structures directly from data, eliminating the need for manually chosen smoothing parameters \cite{lindeberg1998feature}. Multi-scale representations have had a profound impact on computer vision and shape analysis. For example, Lowe's scale-invariant feature transform (SIFT) demonstrated that robust image features can be identified as extrema within scale-space representations \cite{lowe2004distinctive}. Similarly, Mokhtarian and Mackworth developed a curvature-based scale-space framework in which meaningful shape features correspond to structures that persist across multiple levels of smoothing \cite{mokhtarian1992theory}. These studies suggest that scale-space representations provide a natural mechanism for extracting invariant geometric information while suppressing noise. In geometric modeling and surface reconstruction, multi-scale analysis has also played an important role in shape fairing and smoothing. Taubin's signal-processing framework demonstrated how geometric surfaces could be filtered across scales while preserving essential structural properties \cite{taubin1995signal}. Such methods illustrate the broader principle that geometric abstraction should be viewed as a hierarchical process rather than a single-stage operation. The proposed framework adopts this perspective by combining scale-space analysis with topology-guided abstraction. Rather than applying a fixed smoothing level, multiple scale-space representations are evaluated simultaneously and integrated through an adaptive error-driven subdivision strategy. Global shape characteristics are extracted from coarse scales, while local engineering features are preserved through fine-scale analysis and Douglas-Peucker simplification \cite{douglas1973algorithms}. From a neuro-symbolic perspective, this process performs an abstraction operation analogous to human cognition, in which coarse structural concepts emerge before progressively refined detail. Consequently, scale-space theory provides the mathematical foundation for the transition from perceptual geometry to symbolic geometric representation within the proposed architecture.

\section{Emulating the Human Mathematician}

The proposed architecture is motivated by the hypothesis that mathematical intelligence emerges through the interaction of perception, abstraction, and symbolic reasoning. Cognitive studies suggest that humans do not construct mathematical knowledge purely through statistical learning. Instead, they exploit structured priors concerning objects, geometry, space, and topology to extract invariant relationships from sensory observations and subsequently transform these relationships into symbolic concepts \cite{spelke2007core,lake2017building}. This view aligns closely with the objectives of Neuro-Symbolic AI, which seeks to integrate data-driven perception with explicit knowledge representation and reasoning \cite{garcez2002neural,bader2005dimensions,garcez2022neurosymbolic}. Inspired by these observations, our framework decomposes mathematical cognition into two complementary stages: \emph{structural discovery} and \emph{symbolic invention}. The first stage identifies invariant geometric structure directly from observations, while the second constructs symbolic mathematical representations that can support reasoning, editing, and generalization. \subsection{Structural Discovery: Extracting Invariants from Observation} Human cognition appears to possess powerful mechanisms for discovering stable structures within noisy sensory environments. Core Knowledge theory suggests that humans reason about objects, boundaries, containment, and spatial relationships using highly structured internal representations \cite{spelke2007core}. Similarly, Lake et al. argue that abstraction arises from identifying reusable structural regularities rather than merely memorizing examples \cite{lake2017building}. Within the proposed architecture, this capability is realized through topology-driven geometric analysis. Given an unstructured point cloud, the system first employs visual foundation models as perceptual front ends for isolating meaningful observations. These observations are subsequently transformed through the Euclidean Distance Transform (EDT) \cite{maurer2003linear} and Medial Axis Transform (MAT) \cite{blum1967transformation}. Together, these operations reveal intrinsic geometric relationships that remain largely invariant under boundary deformation. Importantly, the MAT does not function as a learned representation. Instead, it acts as an analytical mechanism for extracting topological structure directly from geometric data. This distinguishes the proposed approach from learning-based reconstruction systems that encode geometric knowledge implicitly within network parameters \cite{sharma2018csgnet,wu2021deepcad,khan2024cad,liu2024point2cad}. The resulting skeletal representation serves as an explicit structural description of the observed object and forms the basis for subsequent symbolic processing. From a neuro-symbolic perspective, this stage corresponds to the transformation of perceptual observations into invariant symbolic primitives. Rather than learning geometric concepts from examples, the system derives structure directly from first principles. In this sense, topology functions as an abstraction layer connecting perception to reasoning. \subsection{Symbolic Invention: Constructing Mathematical Reality} Observation alone does not constitute mathematical reasoning. Human mathematicians routinely transform imperfect and incomplete observations into idealized mathematical objects possessing properties that may not exist exactly in physical reality. Lines become infinitely thin, circles become perfectly smooth, and geometric relationships become exact. This process may be interpreted as a transition from observed structure to symbolic representation \cite{lake2017building,garcez2015reasoning,besold2017survey}. The second stage of our framework emulates this transition through symbolic geometric synthesis. Following topological abstraction, the extracted skeletal structure is analyzed across multiple scales using scale-space theory \cite{witkin1983scale,lindeberg1994scale,lindeberg1998scale}. Multi-scale analysis enables the system to distinguish persistent structural features from noise and local measurement artefacts. Global organizational characteristics are encoded at coarse scales, while local engineering details are retained through fine-scale representations. The resulting control points form an explicit symbolic description of the underlying geometry. These symbols are subsequently processed through the Implicit Area Spline framework \cite{li20092d}, yielding a continuous analytical representation of the reconstructed object. Unlike neural implicit representations such as DeepSDF \cite{park2019deepsdf}, the generated model is fully interpretable, analytically defined, and directly editable through symbolic geometric parameters. From the perspective of Neuro-Symbolic AI, the spline formulation serves as a reasoning substrate rather than merely a geometric representation. Control points, spline segments, and algebraic constraints constitute explicit symbolic entities that can be manipulated, combined, and analyzed independently of the original observations. The resulting geometry therefore represents not only a reconstruction of the observed object but also a mathematical hypothesis about its idealized structure. \subsection{Toward Machine Mathematical Abstraction} The proposed architecture suggests a broader interpretation of mathematical intelligence. Rather than viewing intelligence solely as statistical prediction, we view it as a process of transforming observations into invariants and invariants into symbolic representations. This progression can be expressed conceptually as
$$ \mathcal{O} \xrightarrow{\text{Perception}} \mathcal{T} \xrightarrow{\text{Abstraction}} \mathcal{S} \xrightarrow{\text{Reasoning}} \mathcal{C} $$

where $\mathcal{O}$ denotes observations, $\mathcal{T}$ denotes topological invariants, $\mathcal{S}$ denotes symbolic representations, and $\mathcal{C}$ denotes mathematical concepts. Under this interpretation, Scan-to-CAD reconstruction becomes a specific instance of a more general cognitive process. The objective is not merely to recover geometry but to construct symbolic models from empirical observations. This aligns with the broader goals of Neuro-Symbolic AI, namely the integration of perception, abstraction, knowledge representation, and reasoning within a unified computational framework \cite{garcez2022neurosymbolic}. The proposed system therefore serves as a concrete demonstration of how topology-driven abstraction may support machine mathematical reasoning and concept formation.

\section{The Implicit Area Spline}

The final stage of the proposed architecture transforms topological abstractions into explicit mathematical representations. Within the neuro-symbolic framework, this stage plays the role of symbolic synthesis, converting geometric observations into analytical expressions that can be manipulated independently of the original sensory data. Unlike learned latent representations, which encode geometric knowledge implicitly within network parameters, the proposed representation remains entirely analytical, interpretable, and mathematically tractable. \subsection{Closed-Form Symbolic Representation}
Classical CAD systems rely predominantly on explicit boundary representations such as B-Reps and NURBS \cite{requicha1980representations,piegl1997nurbs}. However, these representations require complex numerical procedures for trimming, intersection detection, and topology maintenance. Conversely, neural implicit fields \cite{park2019deepsdf} embed geometry within uninterpretable network weights.

To address these limitations, NeuSOGA employs the Implicit Area Spline formulation \cite{li20092d}. Let a planar shape be defined by a counter-clockwise oriented control polygon $V = \{v_1, v_2, \ldots, v_n\}$ with vertices $v_i = (x_i, y_i) \in \mathbb{R}^2$ ($v_{n+1} \equiv v_1$). The spatial region is represented as the zero level-set of an analytical scalar field:
\begin{equation}
F(x,y) = 0.
\end{equation}
Specifically, for each directed edge $e_i = (v_i, v_{i+1})$, the signed algebraic distance component $\phi_i(x,y)$ is formulated via Green's integral theorem over the polygonal boundary:
\begin{equation}
F(x,y) = \sum_{i=1}^{n} \omega_i(x,y) \cdot \Delta_i(x,y),
\label{eq:area_spline_explicit}
\end{equation}
where $\Delta_i(x,y) = (x_i - x)(y_{i+1} - y) - (x_{i+1} - x)(y_i - y)$ denotes the signed triangular area spanned by $(x,y)$ and edge $e_i$, and $\omega_i(x,y)$ represents localized $C^n$-smooth algebraic blending weights \cite{li20092d}. 

The resulting Area Spline guarantees:
\begin{itemize}
    \item \textbf{Exact Closed-Form Evaluation:} $F(x,y)$ can be constructed to achieve arbitrary-order smoothness, being piecewise polynomial and globally $C^n$-continuous for any prescribed $n$.
    \item \textbf{Analytical Gradient \& Curvature:} First and higher-order derivatives $\nabla F(x,y)$ and $\mathcal{H}(F)$ are available in closed form without finite differences.
    \item \textbf{Direct Symbolic Editability:} Shape deformation is achieved by directly perturbing vertex coordinates $\{v_i\}$ in Eq.~(\ref{eq:area_spline_explicit}).
\end{itemize} 

Unlike numerical approximations or learned implicit functions, the Area Spline is defined entirely through closed-form analytical expressions. The resulting representation is directly evaluable, differentiable, and symbolically manipulable. Furthermore, Li and Tian proved that the Piecewise Algebraic Spline framework supports arbitrary-order continuity, yielding implicit representations with guaranteed $C^n$ smoothness while remaining completely analytical \cite{li20092d}. Consequently, the Area Spline combines four properties that rarely coexist within a single geometric representation: \begin{itemize} \item closed-form analytical evaluation, \item arbitrary-order continuity ($C^n$), \item explicit geometric editability through control points, \item implicit topological robustness. \end{itemize} Within the proposed neuro-symbolic architecture, the control points, spline segments, and algebraic constraints constitute explicit symbolic entities that encode geometric knowledge. Geometric reasoning is therefore performed through analytical manipulation of symbolic expressions rather than adaptation of learned model parameters. \subsection{Additivity and Symbolic Composition} The defining mathematical property of Area Splines is their strict additivity. Let a polygon $P$ be partitioned into a collection of sub-polygons \begin{equation} P=\bigcup_{i=1}^{m}P_i . \end{equation} The Area Spline field satisfies \begin{equation} F_P(x,y) = \sum_{i=1}^{m} F_{P_i}(x,y). \label{eq:additivity} \end{equation} This property follows directly from the additive nature of the underlying area formulation \cite{li20092d,braden1986surveyor}. Shared internal boundaries contribute equal and opposite signed terms, causing their influence to cancel exactly. Consequently, only the external boundary contributes to the final field representation. From a geometric perspective, Eq.~(\ref{eq:additivity}) implies that shape composition can be performed through simple analytical accumulation rather than explicit reconstruction of boundary topology. Unlike conventional boundary representations, the mathematical description of an object is invariant with respect to how it has been partitioned during modeling. This property is of particular significance because it introduces a compositional structure analogous to symbolic reasoning. Complex geometric configurations can be constructed by combining simpler analytical components while preserving mathematical consistency. The representation therefore supports compositional operations directly at the symbolic level, a capability widely regarded as a defining characteristic of human abstraction and reasoning \cite{lake2017building,garcez2022neurosymbolic}. \subsection{Implications for Constructive Solid Geometry} Constructive Solid Geometry (CSG) traditionally represents objects through Boolean operations applied to explicit geometric primitives \cite{requicha1980representations}. Although elegant in theory, practical implementations often require numerically intensive procedures for intersection detection, trimming, reparameterization, and topology maintenance. These operations become increasingly fragile as geometric complexity increases. The Area Spline formulation provides an alternative viewpoint. Since objects are represented as analytical scalar fields, geometric composition can be performed directly through field operations. Internal boundaries vanish automatically through symbolic cancellation, eliminating the need for many of the topology-repair procedures commonly required by explicit geometric representations. Consequently, Boolean composition is transformed from a topology-management problem into an algebraic operation on symbolic field expressions. This significantly reduces geometric complexity while preserving analytical consistency. The resulting representation remains smooth, continuous, and topologically coherent throughout the modeling process. \subsection{Area Splines as a Neuro-Symbolic Representation} Within the broader cognitive architecture proposed in this paper, the Area Spline serves as the final symbolic representation layer. Topological abstraction extracts invariant structural descriptions from sensory observations, while the spline formulation projects these structures into an explicit mathematical domain. This process closely mirrors a fundamental aspect of mathematical cognition. Human mathematicians routinely transform imperfect observations into idealized symbolic objects that can be manipulated independently of the physical world. Lines become infinitely thin, curves become perfectly smooth, and geometric relationships become analytically exact. Similarly, the proposed system transforms noisy point-cloud observations into topological invariants and subsequently into closed-form algebraic representations. The resulting model is therefore more than a geometric reconstruction. It constitutes a symbolic mathematical hypothesis describing the underlying structure of the observed object. In this sense, the Area Spline provides the mechanism through which geometric perception is converted into mathematical representation, completing the transition from observation to abstraction envisioned by the proposed neuro-symbolic framework.

It is important to distinguish the proposed symbolic representation from conventional parametric spline representations such as Bézier curves, B-splines, or NURBS. Parametric splines describe geometric boundaries through a mapping \[ \mathbf{C}(t)=(x(t),y(t)), \] where geometry is recovered by traversing a parameter domain. In contrast, the Area Spline represents the object directly as an implicit field \[ F(x,y)=0, \] thereby encoding the spatial region itself rather than merely its boundary. This distinction is particularly significant for symbolic reasoning because spatial relationships, containment, composition, and topological structure become intrinsic properties of the representation. Furthermore, the additive property of Area Splines enables geometric composition to be expressed through algebraic field operations, providing a form of compositional symbolic representation that is generally not available in conventional parametric spline formulations.

\section{Methodology: A Neuro-Symbolic Pipeline for Geometric Abstraction} 

The NeuSOGA architecture is designed as a computational model of mathematical abstraction. Rather than learning a direct mapping from observations to CAD parameters, the system progressively transforms observations into symbolic mathematical representations through four stages: topological abstraction, perceptual isolation, geometric abstraction, and symbolic synthesis. Formally, the architecture implements the transformation \begin{equation} \mathcal{O} \rightarrow \mathcal{T} \rightarrow \mathcal{G} \rightarrow \mathcal{S}, \end{equation} where \begin{itemize} \item $\mathcal{O}$ denotes observational data, \item $\mathcal{T}$ denotes topological abstractions, \item $\mathcal{G}$ denotes geometric abstractions, \item $\mathcal{S}$ denotes symbolic mathematical representations. \end{itemize} Unlike conventional learning-based Scan-to-CAD systems that encode geometric knowledge within neural network parameters, the proposed framework derives structural representations explicitly from topological and geometric principles before projecting them into an analytical symbolic representation. 

\subsection{Stage 1: Topological Abstraction from Observation} Given an observed point cloud, the system first generates a two-dimensional projection onto a principal viewing plane. Let the resulting occupancy mask be represented as \begin{equation} \mathcal{M}(x,y). \end{equation} To recover intrinsic structural information, an exact Euclidean Distance Transform (EDT) is computed: \begin{equation} D(p) = \min_{q\in\partial \mathcal{M}} \|p-q\|_2, \end{equation} where $\partial \mathcal{M}$ denotes the boundary of the projected shape \cite{maurer2003linear}. The local maxima of the distance field, \begin{equation} P_{\mathrm{core}} = \{p_1,p_2,\ldots,p_n\}, \end{equation} correspond to centres of maximal inscribed regions and provide a topology-aware description of the object's structural organization. Inspired by the principles underlying Blum's Medial Axis Transform \cite{blum1967transformation}, these points function as topological primitives that summarize the object's intrinsic spatial structure. Unlike conventional object detectors, these topological abstractions are derived analytically and require no task-specific training. 

\subsection{Stage 2: Topology-Guided Perceptual Isolation} The discovered structural cores are subsequently used to guide foundation-model perception. For each topological primitive $p_i$, a segmentation is generated using Segment Anything (SAM/SAM2): \begin{equation} \mathcal{P}_i = \mathcal{F}_{SAM} (\mathcal{M},p_i), \end{equation} where $\mathcal{F}_{SAM}$ denotes the segmentation operator \cite{kirillov2023segment,ravi2024sam2}. This differs fundamentally from conventional prompting strategies. Rather than relying on user interaction or arbitrary seeds, segmentation prompts are generated automatically from topological analysis. Consequently, topology guides perception rather than perception discovering topology. The output of this stage is a collection of coherent object components \begin{equation} \mathcal{P} = \{\mathcal{P}_1,\mathcal{P}_2,\ldots,\mathcal{P}_m\}, \end{equation} which form the basis for subsequent geometric abstraction. From a neuro-symbolic perspective, this stage combines neural perception with deterministic geometric reasoning while maintaining a clear separation between the two processes. 

\subsection{Stage 3: Adaptive Multi-Scale Geometric Abstraction} For each segmented component, the boundary contour is analysed across multiple scales using Gaussian scale-space theory \cite{witkin1983scale,lindeberg1994scale,lindeberg1998scale}. Two complementary contour representations are generated: \begin{equation} C_{\mathrm{coarse}}(t) = G_{\sigma_c} * C(t), \end{equation} \begin{equation} C_{\mathrm{fine}}(t) = G_{\sigma_f} * C(t), \end{equation} where $\sigma_c=6.0$ and $\sigma_f=1.5$ in the current implementation. The coarse representation captures the dominant global structure of the object, while the fine representation retains local geometric detail. Their discrepancy is measured as \begin{equation} \epsilon(t) = \| C_{\mathrm{coarse}}(t) - C_{\mathrm{fine}}(t) \|_2 . \end{equation} Regions exhibiting large deviations indicate locations where significant geometric detail would be lost through excessive smoothing. An importance mask is therefore defined as \begin{equation} M(t) = \begin{cases} 1, & \epsilon(t)>\tau,\\ 0, & \text{otherwise}, \end{cases} \end{equation} where $\tau$ is a deviation threshold. To ensure spatial coherence, the binary importance mask is smoothed using a Gaussian kernel, producing a continuous blending function \begin{equation} W(t) = G_{\sigma_w} * M(t). \end{equation} The final hybrid contour is constructed by adaptive feature fusion: \begin{equation} C_{\mathrm{hybrid}}(t) = (1-W(t)) C_{\mathrm{coarse}}(t) + W(t) C_{\mathrm{fine}}(t). \end{equation} Consequently, coarse-scale geometry is preserved throughout most of the contour, while fine-scale structure is selectively reintroduced only in regions containing significant geometric detail. This produces a contour that simultaneously preserves global smoothness and local fidelity. Finally, the hybrid contour is compressed using a Douglas-Peucker polygonal approximation \cite{douglas1973algorithms}, \begin{equation} V= \{v_1,v_2,\ldots,v_k\}, \end{equation} yielding a sparse control polygon that serves as the symbolic geometric abstraction of the original observation.

\subsection{Stage 4: Symbolic Mathematical Synthesis} The final stage projects geometric abstractions into an explicit symbolic mathematical representation. For each control polygon $V_i$, an implicit Area Spline representation is generated according to the Piecewise Algebraic Spline formulation of Li and Tian \cite{li20092d}. The resulting geometry is represented analytically as \begin{equation} F_i(x,y)=0. \end{equation} The complete object is then synthesized through additive field composition \begin{equation} F(x,y) = \sum_{i=1}^{m} F_i(x,y). \end{equation} Because the Area Spline formulation satisfies strict additivity, composite shapes can be constructed through algebraic field accumulation without explicit boundary merging or topological repair \cite{li20092d}. Unlike neural implicit representations, the resulting symbolic representation possesses explicit mathematical structure, arbitrary-order smoothness ($C^n$ continuity), exact analytical evaluation, and direct editability through control points. The output of the pipeline is therefore not merely a reconstructed shape but an explicit symbolic mathematical model of the observed object. 

\subsection{Interpretation as Mathematical Abstraction} The complete architecture implements a hierarchy of abstraction: \begin{equation} \text{Observation} \rightarrow \text{Topology} \rightarrow \text{Geometry} \rightarrow \text{Symbol}. \end{equation} Raw observations are transformed into topology-aware structural cores, topological structures guide neural perception, perceptual components are compressed into geometric primitives, and geometric primitives are projected into analytical symbolic representations through Area Splines. The central hypothesis of this work is that symbolic representation constitutes the critical bridge between perception and mathematical reasoning. The present framework therefore focuses on the transformation \[ \mathcal{O} \rightarrow \mathcal{T} \rightarrow \mathcal{G} \rightarrow \mathcal{S}, \] establishing a neuro-symbolic mechanism by which observations are converted into explicit mathematical representations suitable for subsequent reasoning, editing, and knowledge generation.

\subsection{Implicit Spline Evaluation and Algebraic Composition}
The extracted control points are ordered counter-clockwise using the surveyor's formula \cite{braden1986surveyor}. Under the assumption that the decomposed semantic parts $\{\mathcal{P}_k\}_{k=1}^K$ form a quasi-disjoint partition of the global geometry (i.e., $\mathrm{int}(\mathcal{P}_i) \cap \mathrm{int}(\mathcal{P}_j) = \emptyset$ with oppositely oriented shared boundaries), the composite implicit field satisfies strict algebraic additivity \cite{li20092d}:
\begin{equation}
\mathcal{U}(x,y) = \sum_{k=1}^K F_k(x,y).
\label{eq:additive_union}
\end{equation}
Along any internal interface shared by adjacent components $\mathcal{P}_i$ and $\mathcal{P}_j$, the coincident edges possess reversed traversal directions, causing the corresponding area terms $\Delta_i(x,y)$ and $\Delta_j(x,y)$ in Eq.~(\ref{eq:area_spline_explicit}) to cancel identically. Consequently, internal boundaries vanish automatically without explicit polygon-clipping algorithms, yielding a seamless global representation $\mathcal{U}(x,y) = 0$. In cases where segmented components exhibit non‑negligible spatial overlap, standard smooth $R$-functions (e.g., $R$-disjunction $\mathcal{R}_\vee(F_1, F_2) = F_1 + F_2 - \sqrt{F_1^2 + F_2^2}$) can be seamlessly applied; however, Li’s shape‑preserving implicit blending operators\cite{Li2007ImplicitBlending} provide more reliable control over boundary behaviour and are therefore preferred in our formulation.

\subsection{Symbolic Representation as the Outcome of Abstraction} The central hypothesis of this work is that symbolic representation constitutes the essential bridge between perception and mathematical reasoning. Modern neural systems are highly effective at extracting information from sensory data, but their knowledge is typically encoded within latent parameters that are not directly accessible to analytical manipulation. In contrast, human mathematical reasoning depends upon explicit symbolic objects such as equations, functions, geometric constructions, and logical statements. Such representations possess semantics that are independent of the observations from which they were derived and can therefore support reasoning, generalization, and knowledge composition. The proposed architecture is designed explicitly to perform the transformation \[ O \rightarrow T \rightarrow G \rightarrow S, \] where observations are converted into topological invariants, topological invariants into geometric abstractions, and geometric abstractions into symbolic mathematical representations. Within this framework, the Implicit Area Spline representation constitutes the symbolic layer. Given a set of control points \[ V=\{v_1,v_2,\ldots,v_n\}, \] the geometry is represented analytically by a scalar field \[ F(x,y)=0. \] Unlike latent neural representations, the resulting symbolic object possesses explicit mathematical structure, supports closed-form evaluation, exhibits arbitrary-order continuity, and satisfies additive composition properties \cite{li20092d}. The output of the proposed system is therefore not merely a geometric reconstruction but an explicit mathematical model whose structure is available for subsequent symbolic manipulation and reasoning.

\section{Experimental Evaluation}
To validate the effectiveness, structural fidelity, and generalizability of NeuSOGA, we conduct systematic evaluations across diverse geometric configurations. Unlike data-driven reconstruction networks that rely heavily on category-specific training distributions, our experiments are designed to evaluate the zero-shot capability of the neuro-symbolic pipeline in transforming raw sensory observations into compact, accurate, and editable analytical representations:
\begin{equation}
\mathcal{O} \rightarrow \mathcal{T} \rightarrow \mathcal{G} \rightarrow \mathcal{S}.
\end{equation}

where observations are converted into topological abstractions, geometric abstractions, and finally symbolic mathematical representations. \subsection{Experimental Setup} Experiments were conducted using point-cloud models from the ModelNet40 dataset, a widely used benchmark for geometric learning and shape analysis. Three representative object categories were selected to evaluate the robustness of the proposed framework across substantially different structural configurations: \begin{itemize} \item Airplane, \item Chair, \item Lamp. \end{itemize} Each object was normalized and projected onto three principal orthographic views: \begin{itemize} \item Top view (XY plane), \item Front view (XZ plane), \item Side view (YZ plane). \end{itemize}

For each projection, the complete neuro-symbolic pipeline was executed: 
\[ \begin{aligned} \text{Projection} &\rightarrow \text{EDT Core Discovery} \rightarrow \text{SAM Segmentation} \\ &\rightarrow \text{Adaptive Abstraction} \rightarrow \text{Area Spline Synthesis} \end{aligned} \]

No category-specific training, CAD templates, geometric priors, or reconstruction supervision were used.
\subsection{Evaluation Criteria} Because the proposed framework is designed to generate symbolic representations rather than merely geometric reconstructions, evaluation focuses on four qualitative criteria: 
\begin{enumerate} 
\item Structural preservation, 
\item Topological consistency, 
\item Representation compactness, 
\item Symbolic expressiveness. 
\end{enumerate} 

Structural preservation evaluates whether major geometric structures remain recognizable following abstraction. Topological consistency evaluates whether the generated implicit representation preserves the connectivity and organization of the original object. Representation compactness evaluates the degree to which complex boundaries can be represented using sparse control polygons. Symbolic expressiveness evaluates whether the final output constitutes an explicit mathematical representation rather than a numerical approximation or latent encoding. 

\begin{figure*}[t] \centering 
\includegraphics[scale=0.82]{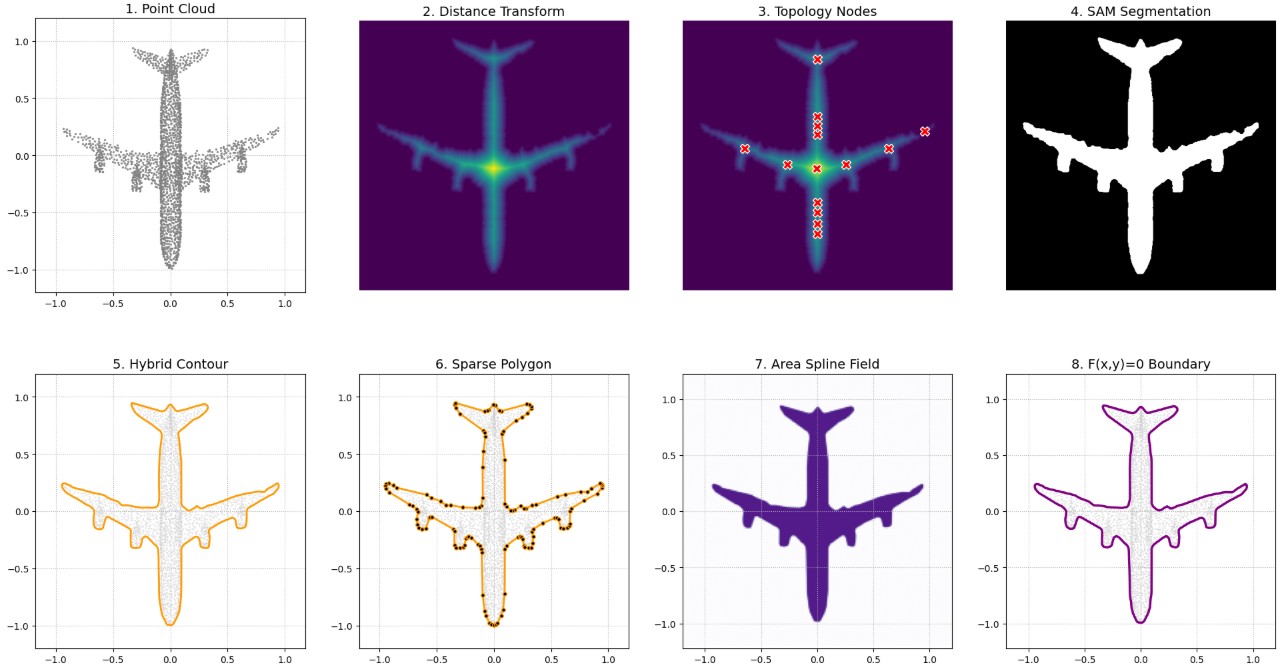}
\includegraphics[scale=0.82]{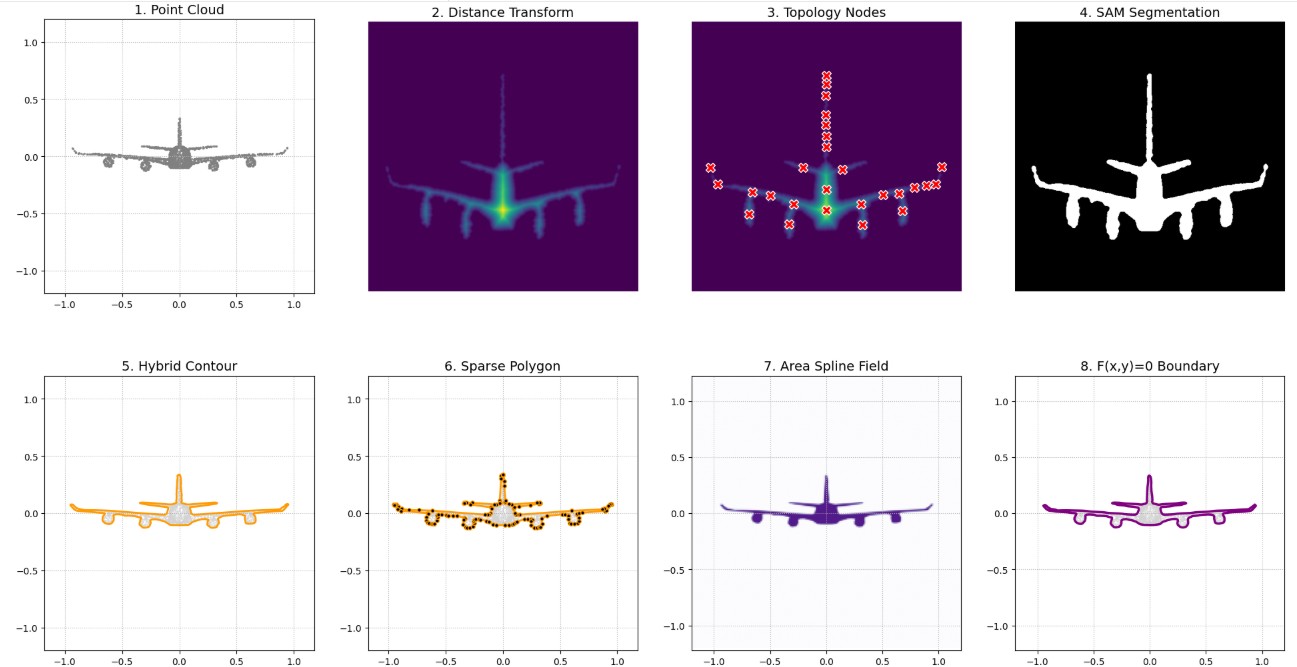}
\includegraphics[scale=0.82]{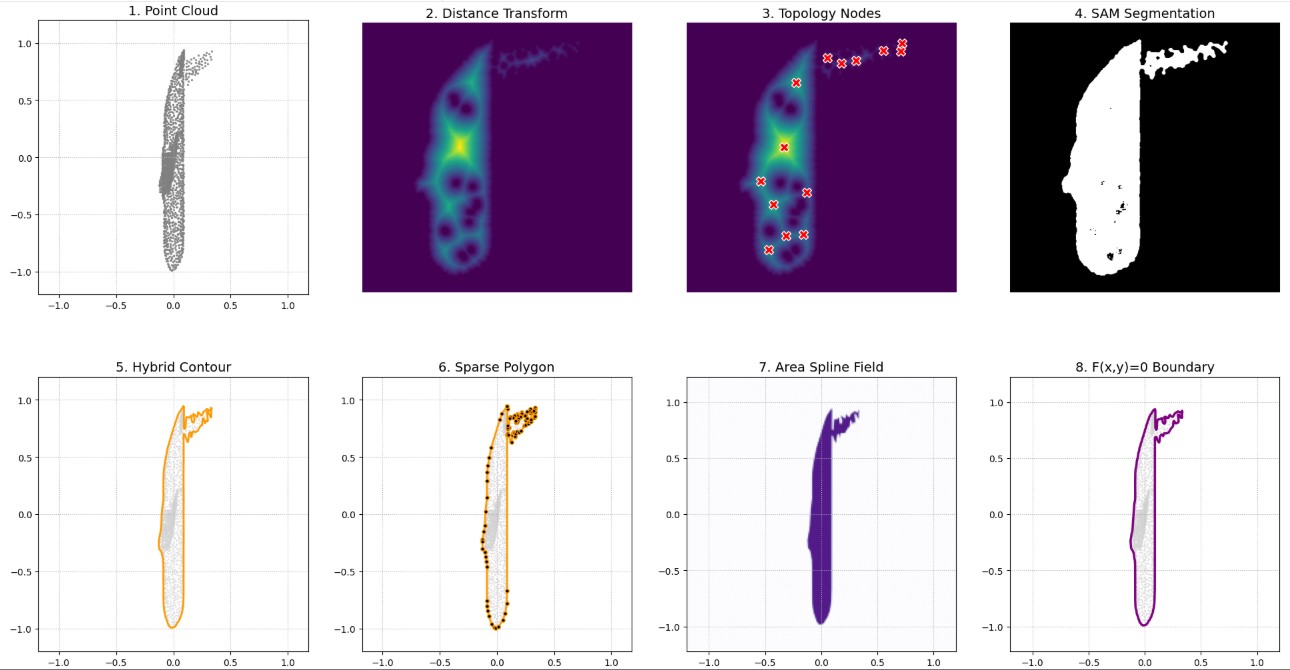}
\caption{ Application of NeuSOGA to an airplane point-cloud model. Rows correspond to the top (XY), front (XZ), and side (YZ) projections of the original 3D object. For each projection, the proposed neuro-symbolic framework progressively transforms the observation into topology-aware structural abstractions, segmented object regions, adaptive geometric abstractions, and finally an explicit symbolic mathematical representation using Implicit Area Splines. The resulting representations preserve the dominant aircraft morphology while significantly reducing geometric complexity, demonstrating the complete transformation from observation to symbol ($\mathcal{O}\rightarrow\mathcal{T}\rightarrow\mathcal{G}\rightarrow\mathcal{S}$). }

\label{fig:airplane_results} \end{figure*}

\subsection{Airplane Reconstruction} Figure~\ref{fig:airplane_results} presents the results obtained for the airplane model. The topological abstraction stage successfully identified the major structural components of the aircraft despite substantial variations between viewing directions. The adaptive coarse-to-fine abstraction mechanism preserved high-curvature features such as wing tips, stabilizers, and engine structures while simultaneously recovering the dominant fuselage geometry. The resulting polygonal abstraction required between 97 and 132 control points depending upon projection direction, representing a substantial reduction in geometric complexity relative to the original point cloud. Despite this reduction, the resulting Area Spline representation accurately preserved the characteristic aircraft morphology. Particularly noteworthy is the preservation of global symmetry in both the top and front views. The symbolic representation reconstructed through the Area Spline formulation produces a smooth analytical boundary while retaining fine-scale structural details introduced through adaptive scale-space refinement. 

\begin{figure*}[t] \centering 
\includegraphics[scale=0.72]{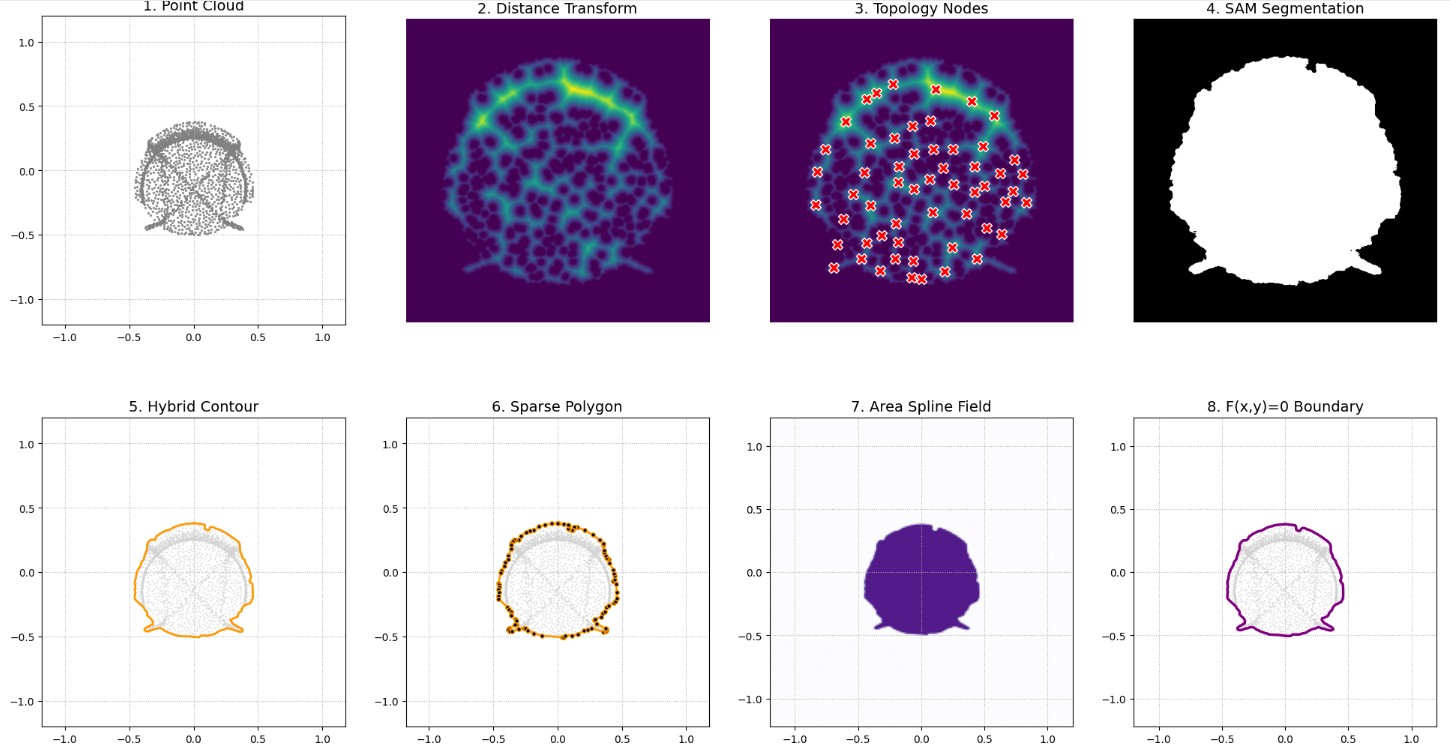} 
\includegraphics[scale=0.8]{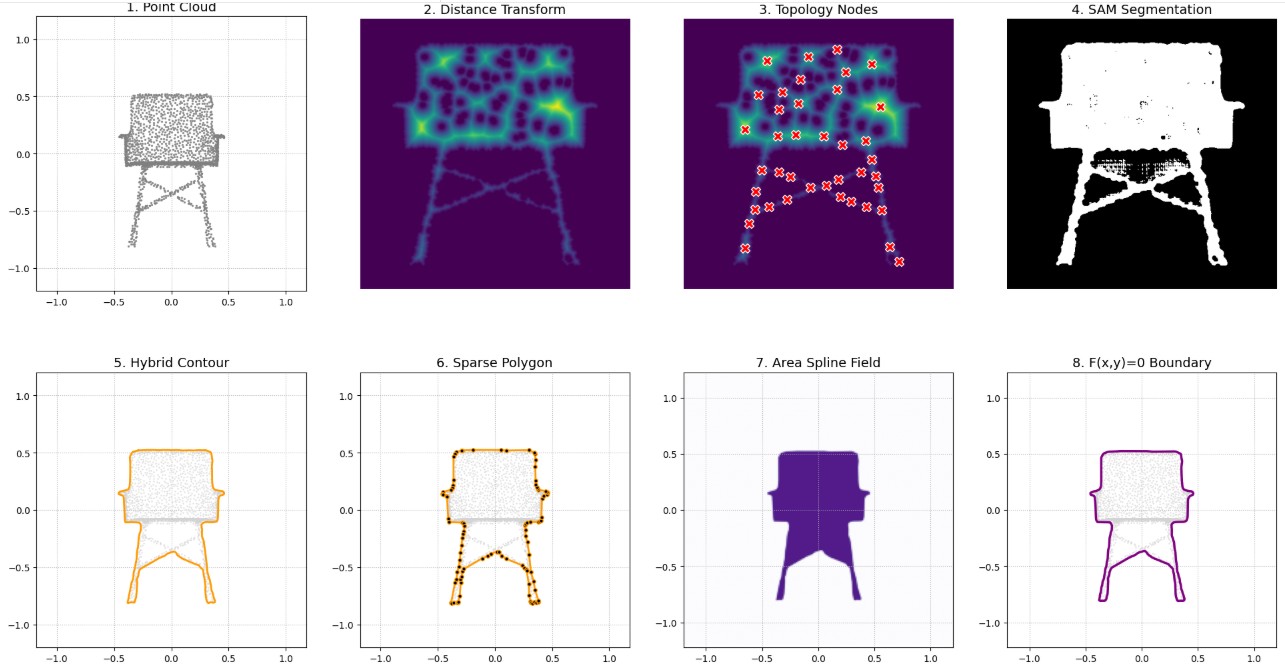} 
\includegraphics[scale=0.72]{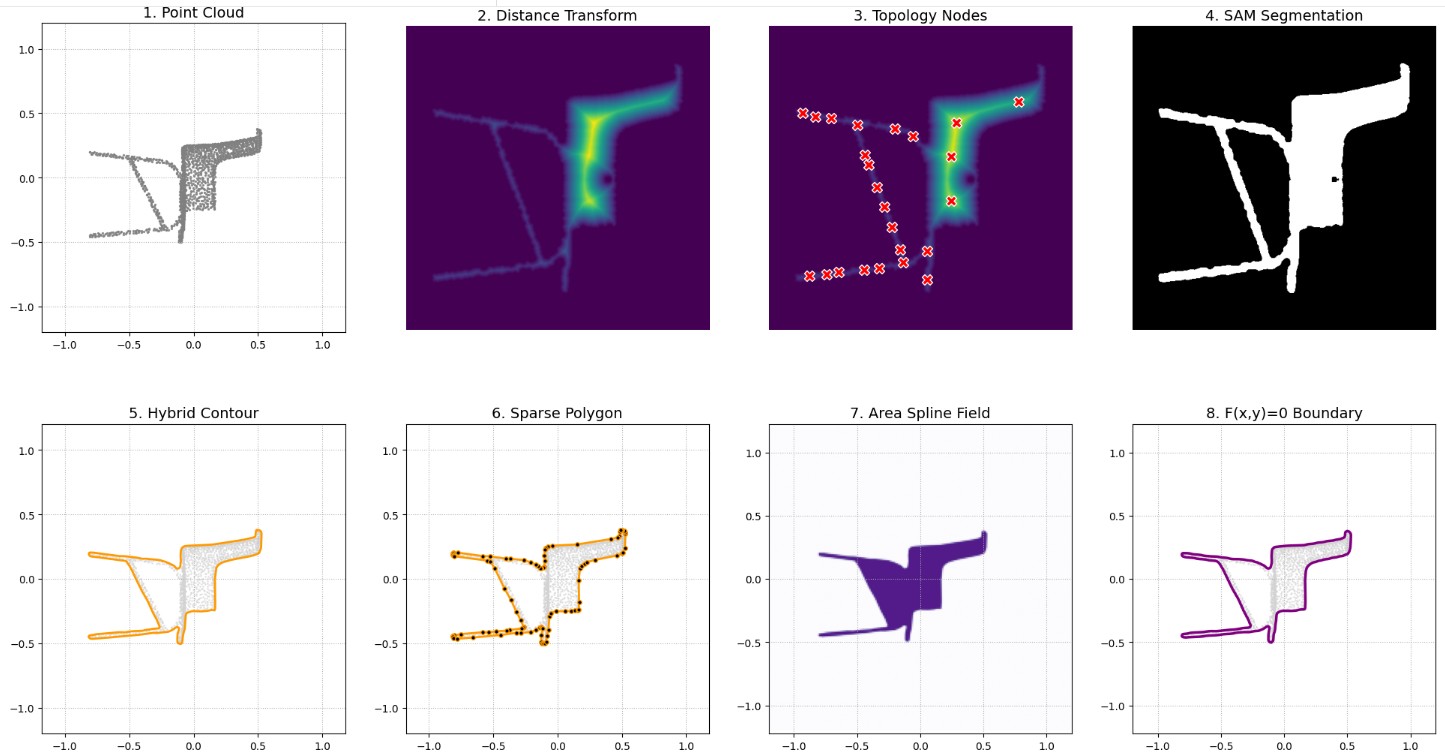} 
\caption{ Application of NeuSOGA to a chair point-cloud model. Rows correspond to the top (XY), front (XZ), and side (YZ) projections. Despite the presence of sharp corners, support structures, and complex connectivity, NeuSOGA successfully extracts topology-aware primitives, performs adaptive geometric abstraction, and synthesizes explicit symbolic representations through Implicit Area Splines. The resulting symbolic models preserve the essential seat, backrest, and support-leg structures while providing a compact analytical representation. }
\label{fig:chair_results} \end{figure*}

\subsection{Chair Reconstruction} Figure~\ref{fig:chair_results} illustrates results for a structurally different object category. Unlike the airplane, the chair contains pronounced rectilinear structures, sharp transitions, and disconnected geometric support elements. These characteristics represent a challenging test for smoothing-based reconstruction methods. The proposed abstraction process successfully preserved the dominant seat, backrest, and support-leg structures. Although substantial geometric simplification was performed, the resulting symbolic representations maintain clear correspondence with the original object topology. The side-view reconstruction is particularly significant because it demonstrates that the adaptive abstraction mechanism preserves engineering-style geometric features without excessive smoothing. The resulting symbolic representation remains interpretable and structurally faithful despite substantial reduction in boundary complexity. 

\begin{figure*}[t] \centering 
\includegraphics[scale=0.8]{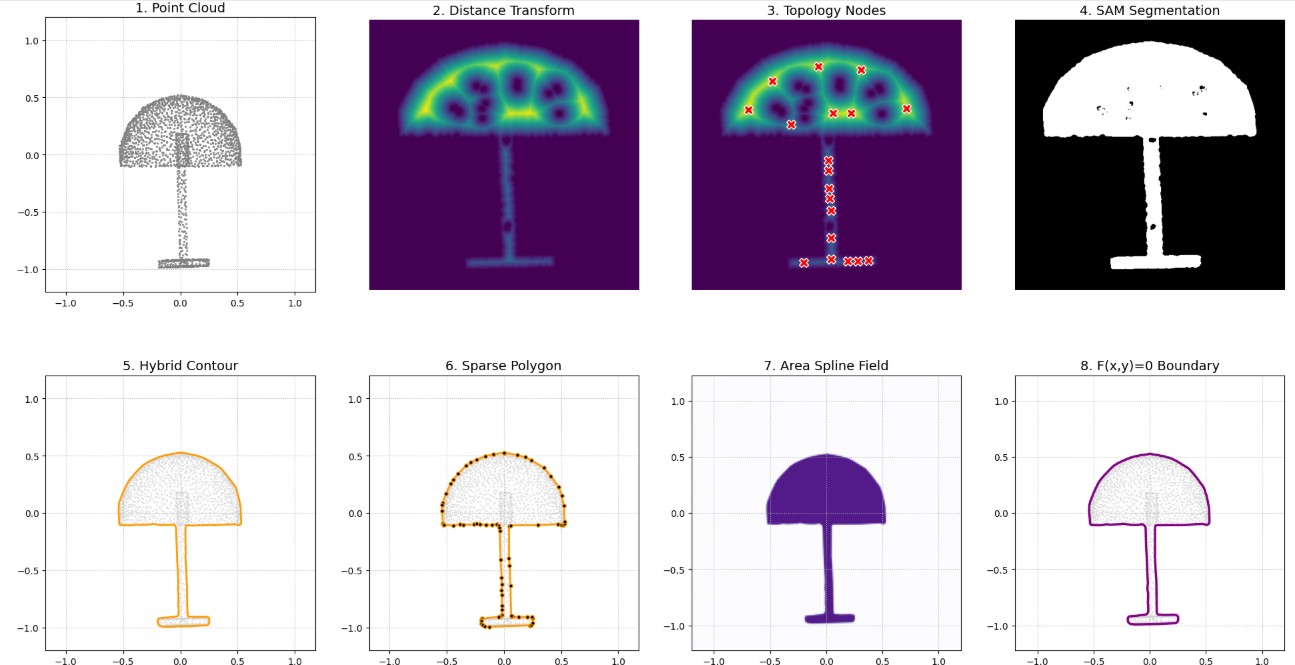} 
\includegraphics[scale=0.8]{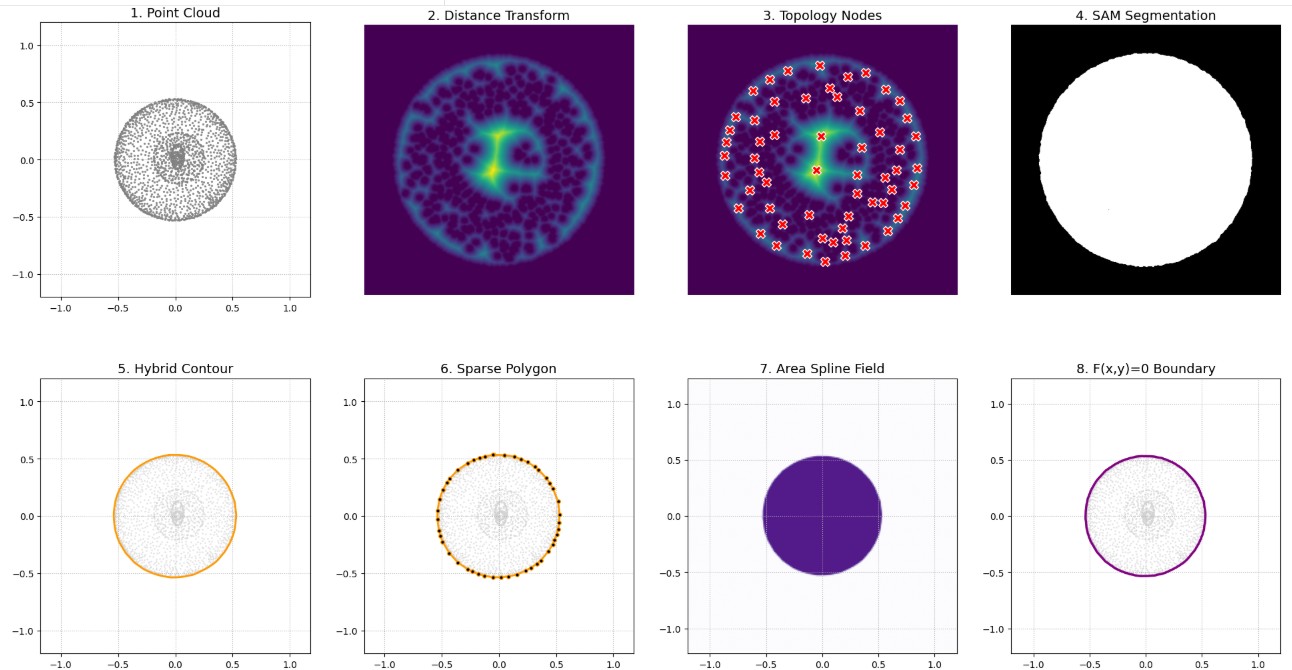} 
\includegraphics[scale=0.8]{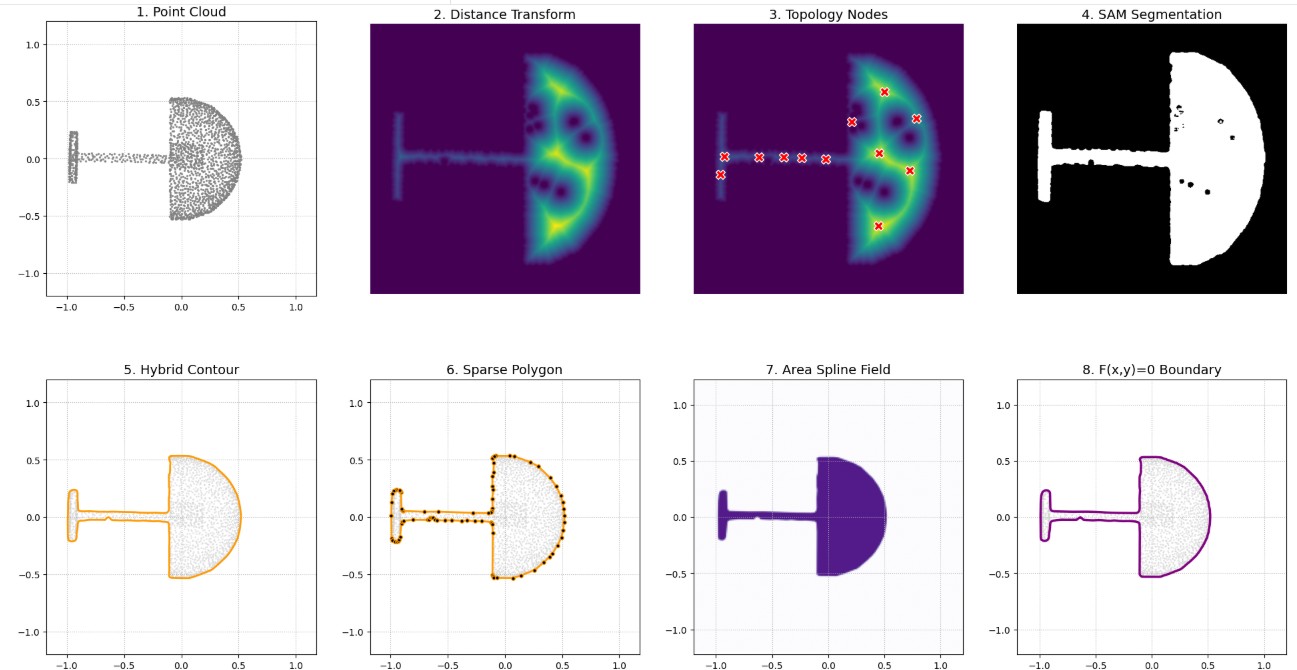} 
\caption{ Application of NeuSOGA to a lamp point-cloud model. Rows correspond to the top (XY), front (XZ), and side (YZ) projections. The proposed pipeline identifies topology-aware structural cores, isolates coherent object regions through foundation-model segmentation, and generates smooth symbolic representations using adaptive abstraction and Implicit Area Splines. The resulting analytical models accurately capture both the global shape and curved geometric features of the lamp, illustrating the robustness of the observation-to-symbol transformation across different object categories. } 
\label{fig:lamp_results} \end{figure*}

\subsection{Lamp Reconstruction} Figure~\ref{fig:lamp_results} presents results for the lamp model. This example exhibits strong rotational symmetry and smooth curved surfaces, providing a complementary test case to the highly articulated airplane and chair geometries. Across all three projections, the proposed framework reconstructed clean implicit boundaries that faithfully captured the dominant geometric features of the lamp. The hemispherical shade, cylindrical support structure, and circular base were all preserved through the abstraction process. The front projection is particularly illustrative because the generated Area Spline representation recovers a nearly perfect circular profile using a sparse symbolic representation. This demonstrates the ability of the framework to transform noisy point-cloud observations into smooth analytical geometric descriptions. 

\subsection{Robustness Across Modalities and Viewing Directions} 

To evaluate the generality of the proposed neuro-symbolic architecture, we investigated its behaviour across both different sensing modalities and different viewing conditions. Specifically, we performed two complementary robustness studies. First, we applied the complete $O\rightarrow T\rightarrow G\rightarrow S$ pipeline to segmented binary observations extracted from the COCO 2017 dataset. Second, we evaluated the framework using non-canonical projections of ModelNet40 point-cloud objects generated along an arbitrary viewing direction $(1,1,1)$. Together, these experiments assess whether the abstraction process depends upon a particular input modality or viewing configuration.

\begin{figure*}[t] \centering 
\includegraphics[scale=0.14]{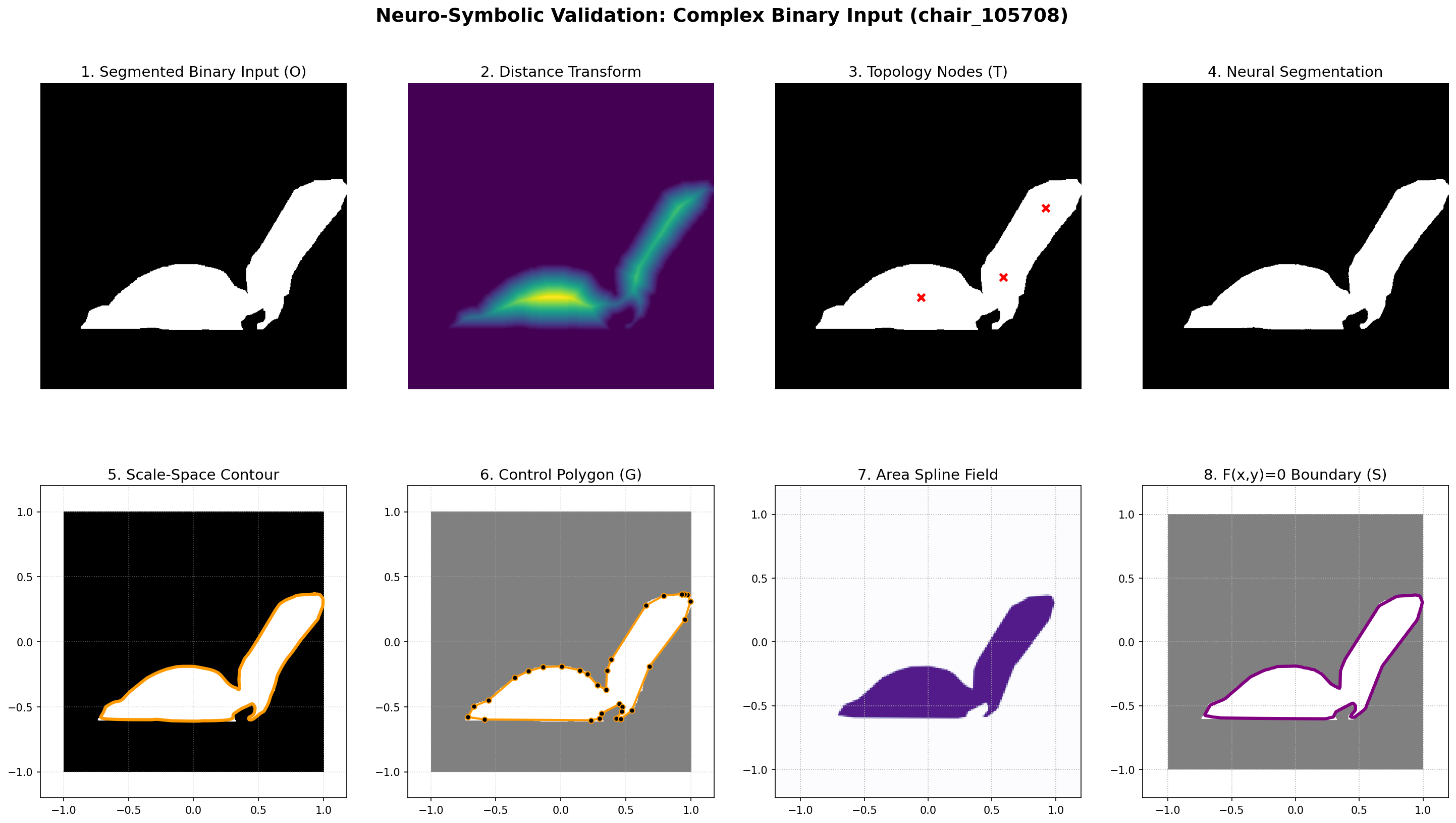} \includegraphics[scale=0.14]{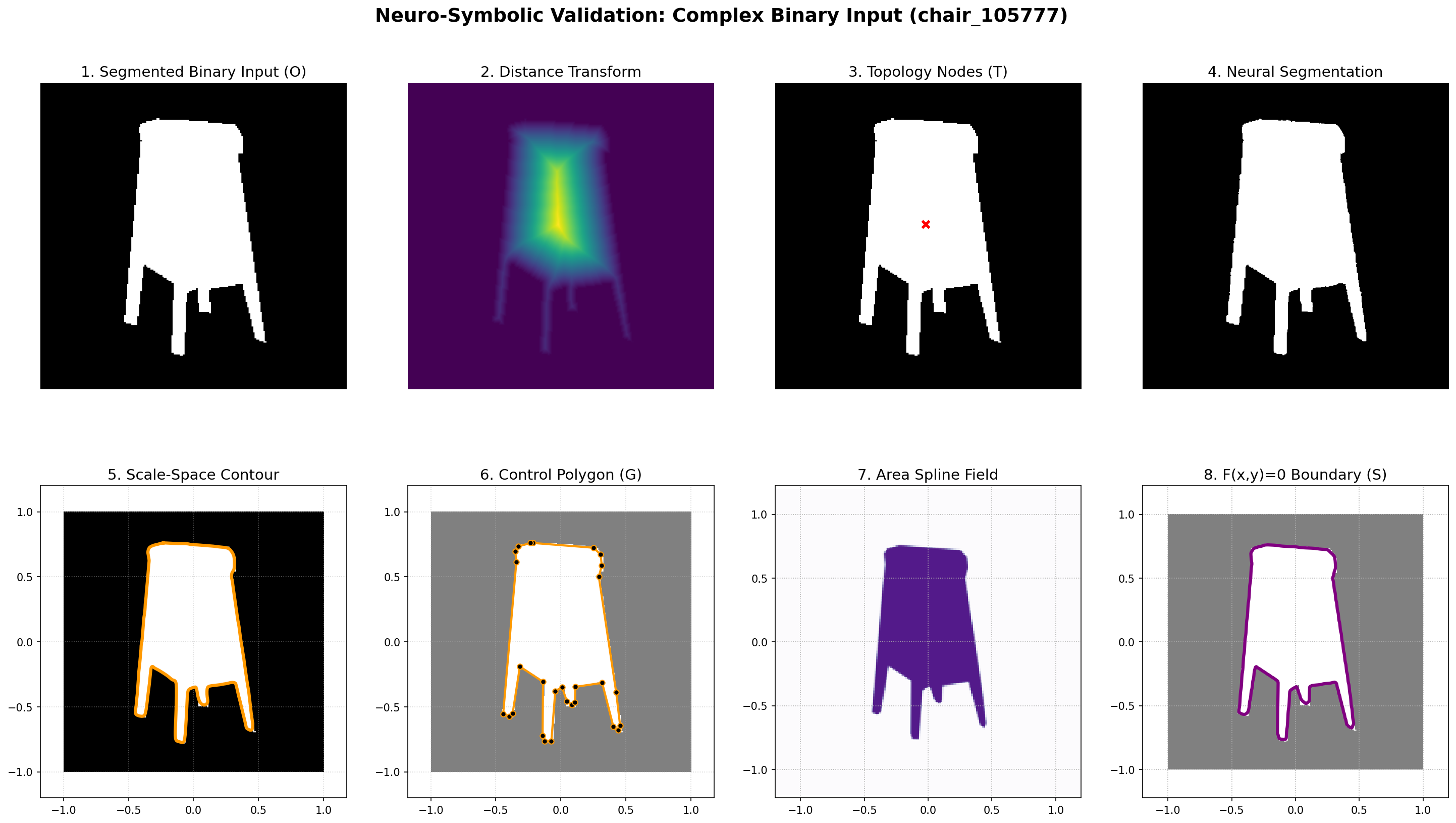}
\includegraphics[scale=0.14]{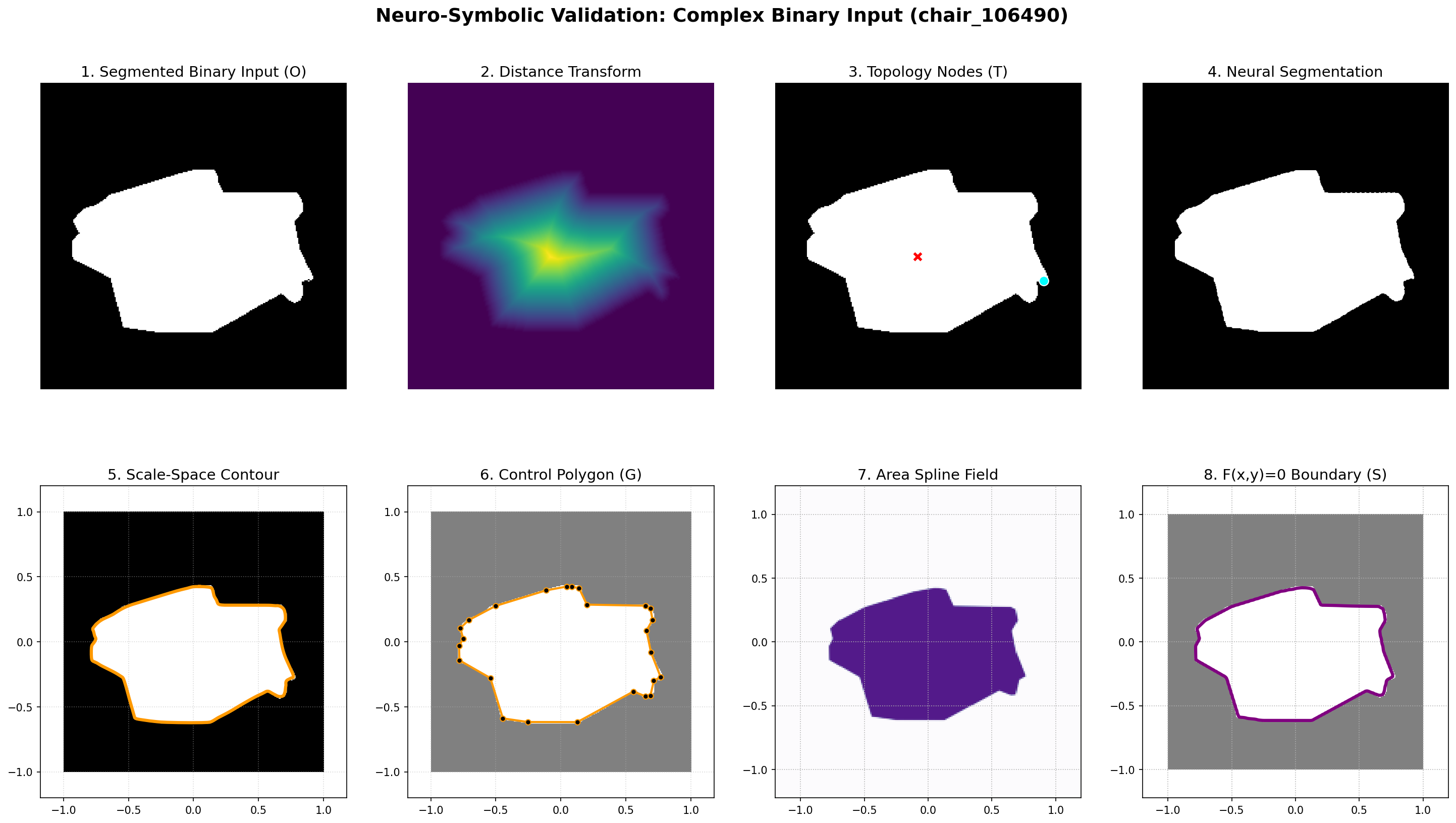} \includegraphics[scale=0.14]{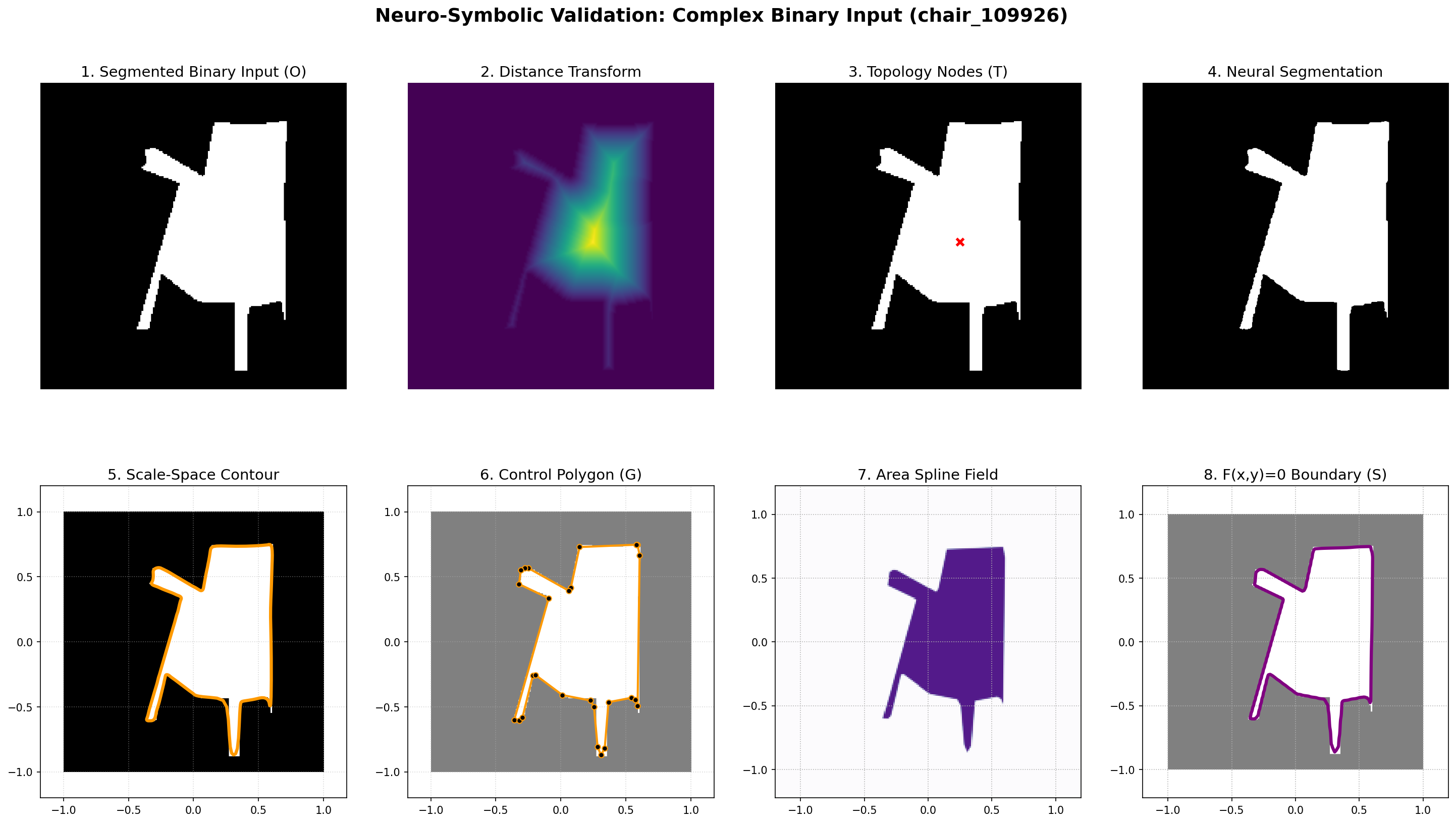}\\
\centering {(a) Chair}\\
\includegraphics[scale=0.14]{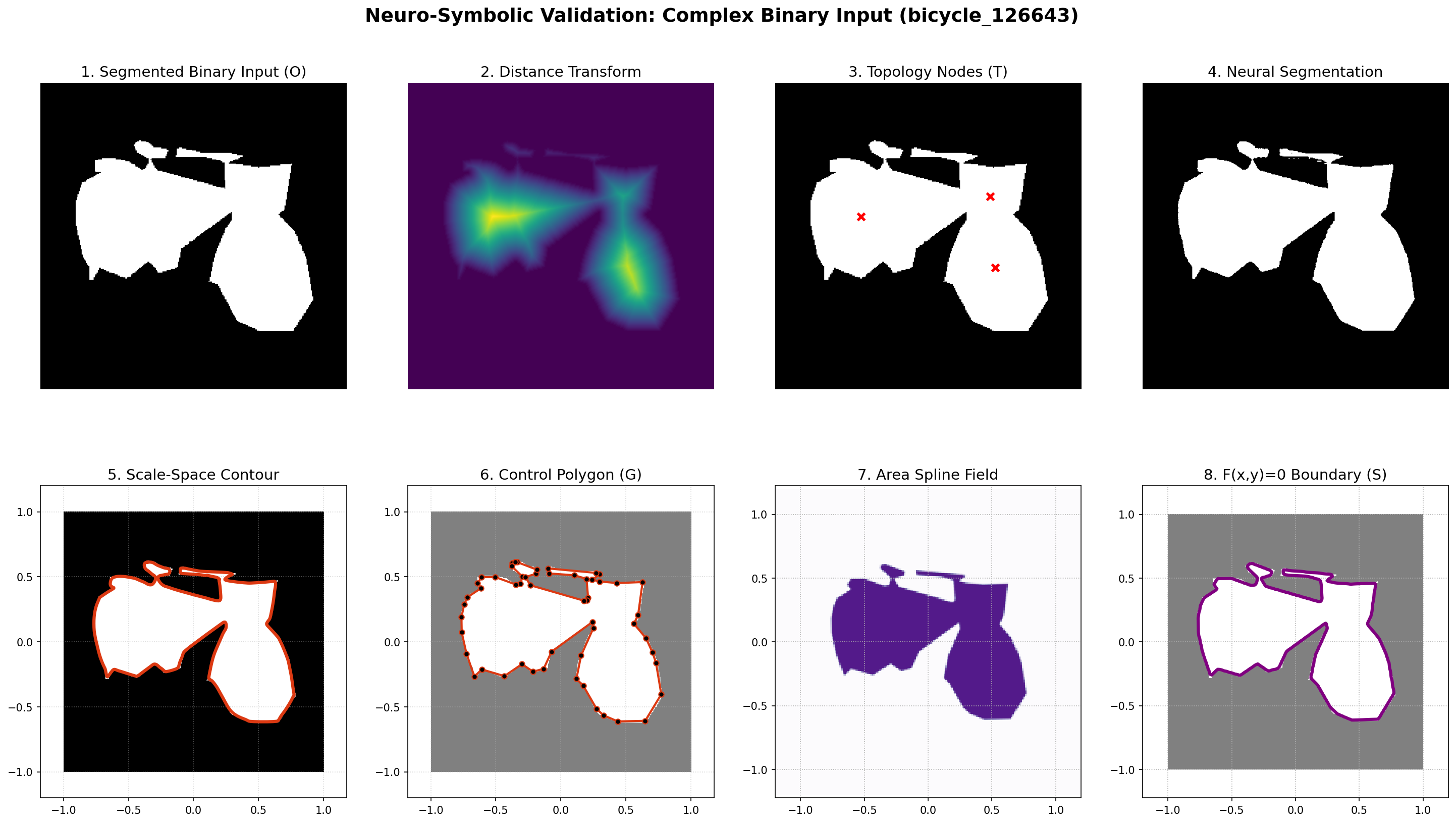} %\includegraphics[scale=0.15]{bicycle_2.png} 
\includegraphics[scale=0.14]{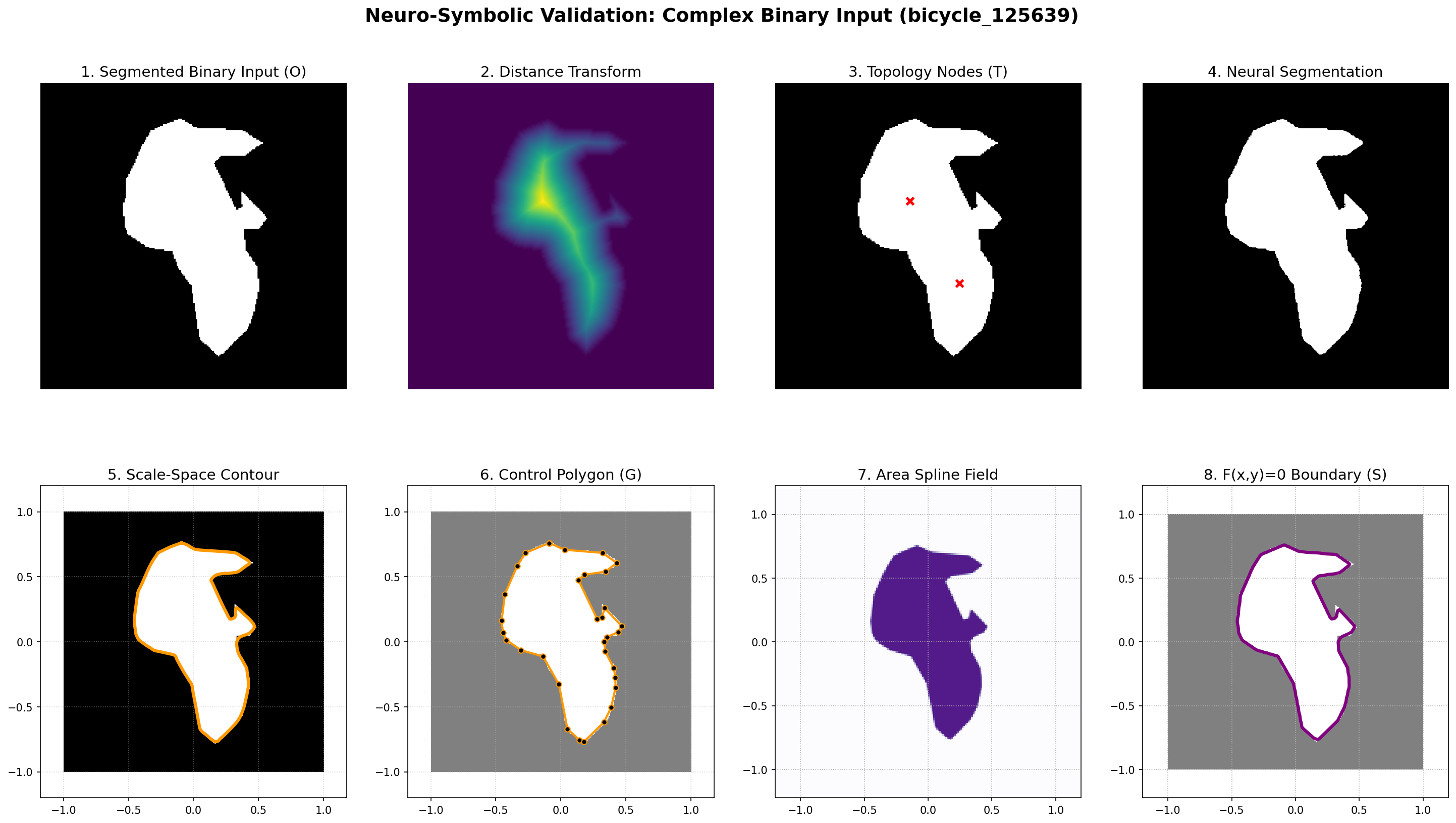} \includegraphics[scale=0.14]{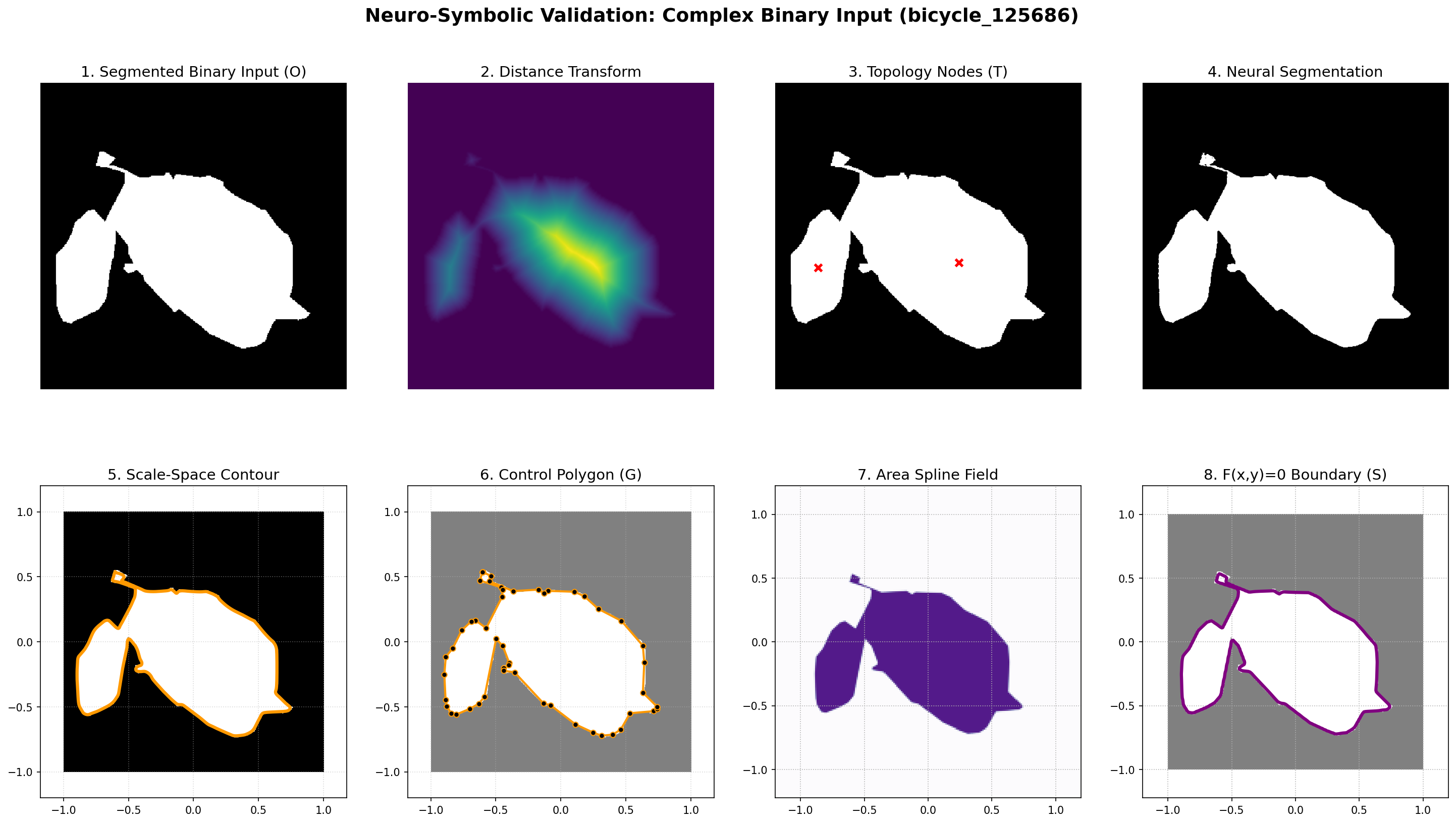}
\includegraphics[scale=0.14]{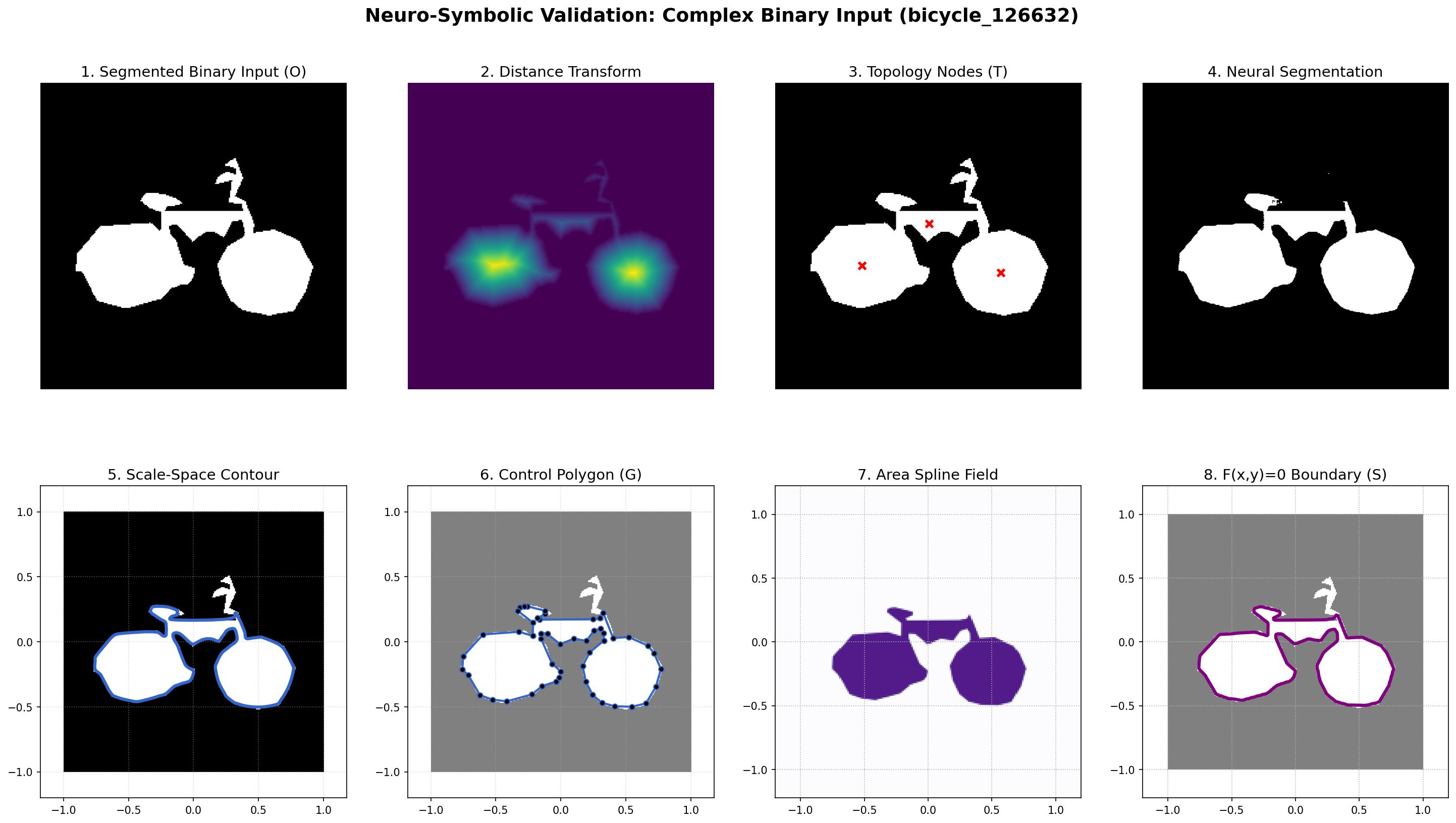} \\
\centering {(b) Cycle} \\

\caption{ Robustness evaluation of the proposed neuro-symbolic abstraction framework on segmented optical observations from the COCO dataset. The upper two rows correspond to a chair object, while the lower two rows correspond to a bicycle object. For each object, the eight columns illustrate the complete observation-to-symbol transformation ($\mathcal{O}\rightarrow\mathcal{T}\rightarrow\mathcal{G}\rightarrow\mathcal{S}$) from left to right: (1) segmented binary observation ($\mathcal{O}$), (2) Euclidean Distance Transform, (3) topology-aware structural cores and internal void nodes ($\mathcal{T}$), (4) topology-guided foundation-model segmentation, (5) adaptive multi-scale contour extraction, (6) sparse geometric control polygon ($\mathcal{G}$), (7) Implicit Area Spline field representation, and (8) the final symbolic mathematical boundary $F(x,y)=0$ ($\mathcal{S}$). The results demonstrate that the proposed architecture preserves complex topological structures, internal voids, and thin mechanical features while generating explicit symbolic representations directly from segmented optical observations. }
\label{fig:Image_results} \end{figure*}

A central objective of the proposed neuro-symbolic architecture is to perform abstraction independently of the sensing modality used to generate the observation. To evaluate this property, we extended the experimental analysis beyond projected point clouds and applied the complete $O \rightarrow T \rightarrow G \rightarrow S$ pipeline to segmented optical observations. Specifically, binary object masks were extracted from the COCO 2017 dataset and injected directly into the Observation layer ($\mathcal{O}$). These masks contain substantially different characteristics from the projected point-cloud observations used in previous experiments. While point clouds are sparse and incomplete, binary masks provide dense occupancy information with sharply defined boundaries. Successful processing of both modalities would therefore indicate that the architecture is responding to abstract spatial structure rather than modality-specific characteristics. Unlike conventional perception pipelines that often require separate processing pathways for different data sources, the proposed framework applies the same abstraction sequence regardless of the underlying representation: \[ \mathcal{O} \rightarrow \mathcal{T} \rightarrow \mathcal{G} \rightarrow \mathcal{S}. \] No modality-specific modifications were introduced during evaluation. For binary observations, the Euclidean Distance Transform first generates a spatial distance field from which topology-aware structural cores and internal void nodes are extracted. These topological abstractions subsequently guide foundation-model perception and contour isolation. Adaptive multi-scale abstraction then converts the resulting contours into sparse geometric control polygons, which are finally synthesized into symbolic mathematical representations through the Area Spline formulation. An interesting observation emerges when applying the framework to clean binary masks. Because the object has already been segmented, the foundation-model layer frequently behaves as a near-identity mapping. While this may appear computationally redundant in idealized settings, retaining the perceptual layer provides several important advantages. First, it preserves a modality-independent architecture. Point clouds, segmented images, occupancy grids, medical images, and other spatial observations can be processed using the same computational pathway without introducing heuristic branching logic. Second, the perceptual layer provides a mechanism for handling imperfect observations. Real-world binary masks often contain segmentation artifacts, disconnected regions, missing boundaries, aliasing effects, and spurious noise. Foundation-model perception offers an additional abstraction layer capable of improving structural consistency before symbolic processing. Third, topology-guided prompting introduces a level of semantic awareness that is not available through conventional contour extraction alone. Multiple structural cores and internal void nodes can be used to distinguish disconnected regions, preserve internal cavities, and guide segmentation of complex overlapping structures. Figure~\ref{fig:Image_results} illustrates representative results obtained from bicycle and chair objects containing multiple internal voids and thin structural features. Despite substantial topological complexity, NeuSOGA successfully identifies both positive structural regions and negative interior spaces. The resulting control polygons capture the dominant geometric organization of the objects, while the symbolic Area Spline representation preserves fine structural details such as bicycle-frame openings, wheel cavities, and chair support gaps. Most importantly, the experiments demonstrate that the symbolic representation layer remains unchanged across modalities. Whether the input originates from a projected point cloud or a segmented optical observation, the final output consists of an explicit analytical representation \[ F(x,y)=0, \] possessing arbitrary-order smoothness, additive composition, and direct symbolic interpretability. This result provides evidence that the proposed architecture is learning neither object categories nor sensing modalities. Rather, it operates on topological and geometric abstractions that remain invariant across different forms of observation. Consequently, the experiments support the broader hypothesis that symbolic representation formation can be achieved through a modality-independent abstraction process, reinforcing the role of the proposed architecture as a general neuro-symbolic framework rather than a domain-specific reconstruction pipeline.

In addition to modality generalization, we investigated viewpoint robustness using arbitrary projections of three-dimensional point-cloud models from the ModelNet40 dataset. Whereas previous experiments focused on canonical orthographic views (top, front, and side), the additional evaluation employed projections along the $(1,1,1)$ viewing direction. This configuration simultaneously exposes multiple object surfaces and typically produces more complex contour structures than standard orthographic projections. The resulting observations represent a particularly challenging test case because self-occlusion, overlapping structural components, and non-axis-aligned features generate significantly more intricate topological and geometric configurations. Despite these challenges, NeuSOGA successfully identified topology-aware structural cores, extracted meaningful geometric abstractions, and synthesized continuous Area Spline representations. The resulting symbolic models preserved the dominant structural organisation of the original objects while maintaining compact analytical descriptions. These experiments demonstrate that the transformation \[ \mathcal{O} \rightarrow \mathcal{T} \rightarrow \mathcal{G} \rightarrow \mathcal{S} \] is not dependent upon canonical viewpoints. Instead, the abstraction process operates upon structural and topological information that remains largely invariant across viewing directions, further supporting the hypothesis that the proposed architecture learns abstract geometric organisation rather than view-specific representations.

\begin{figure*}[t] \centering 

\includegraphics[scale=0.14]{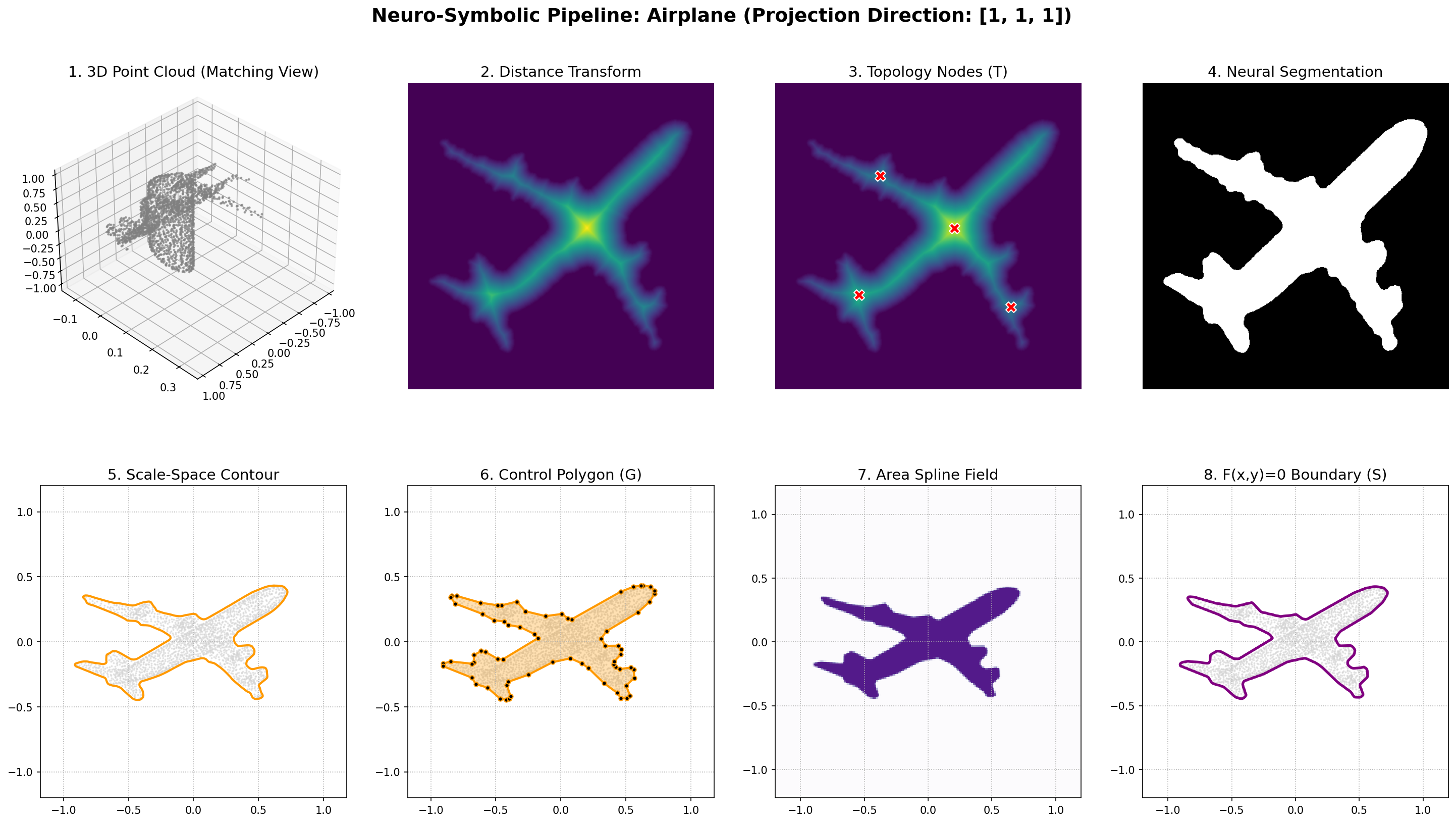} \includegraphics[scale=0.14]{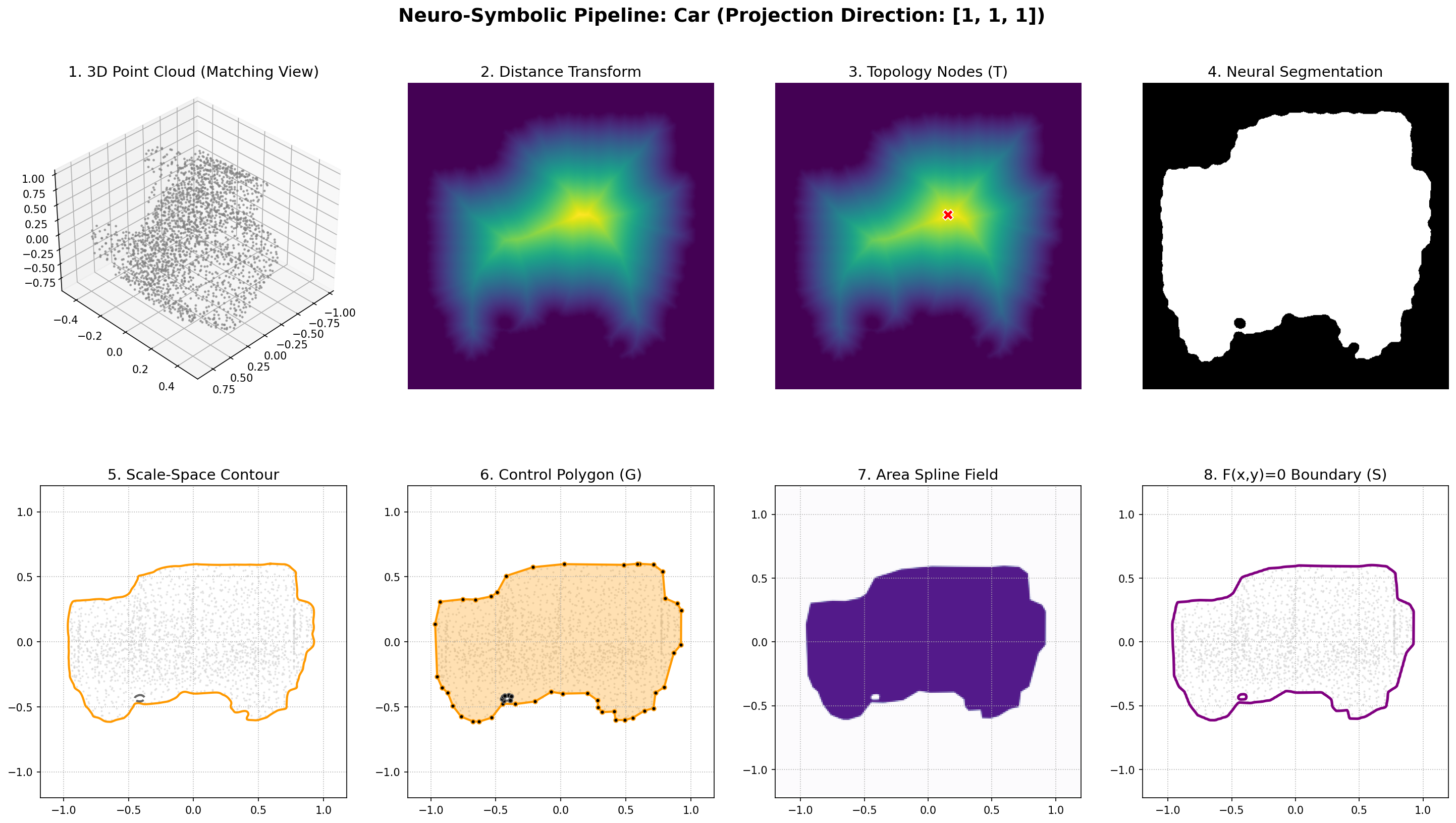}
\includegraphics[scale=0.14]{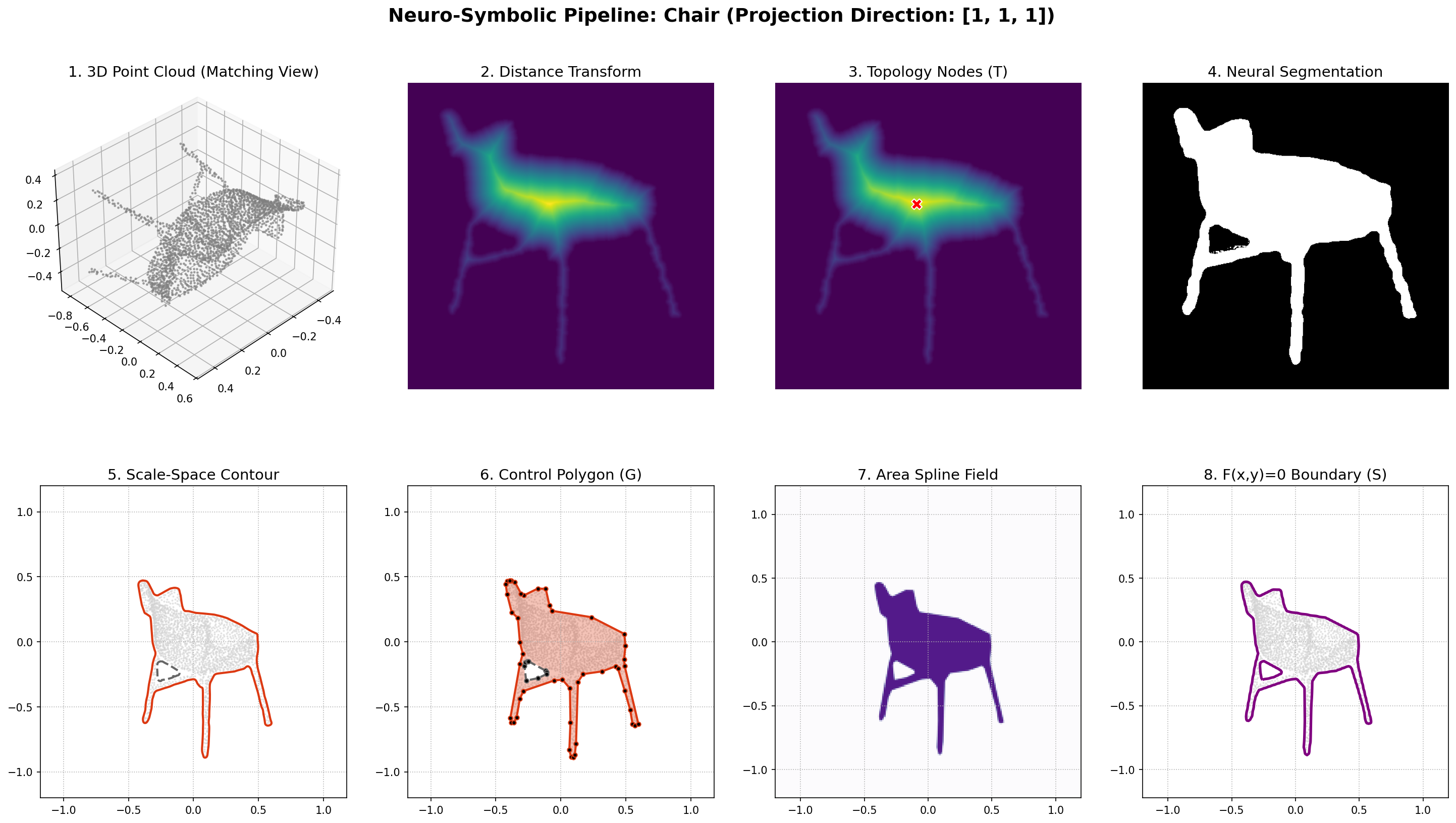} \includegraphics[scale=0.14]{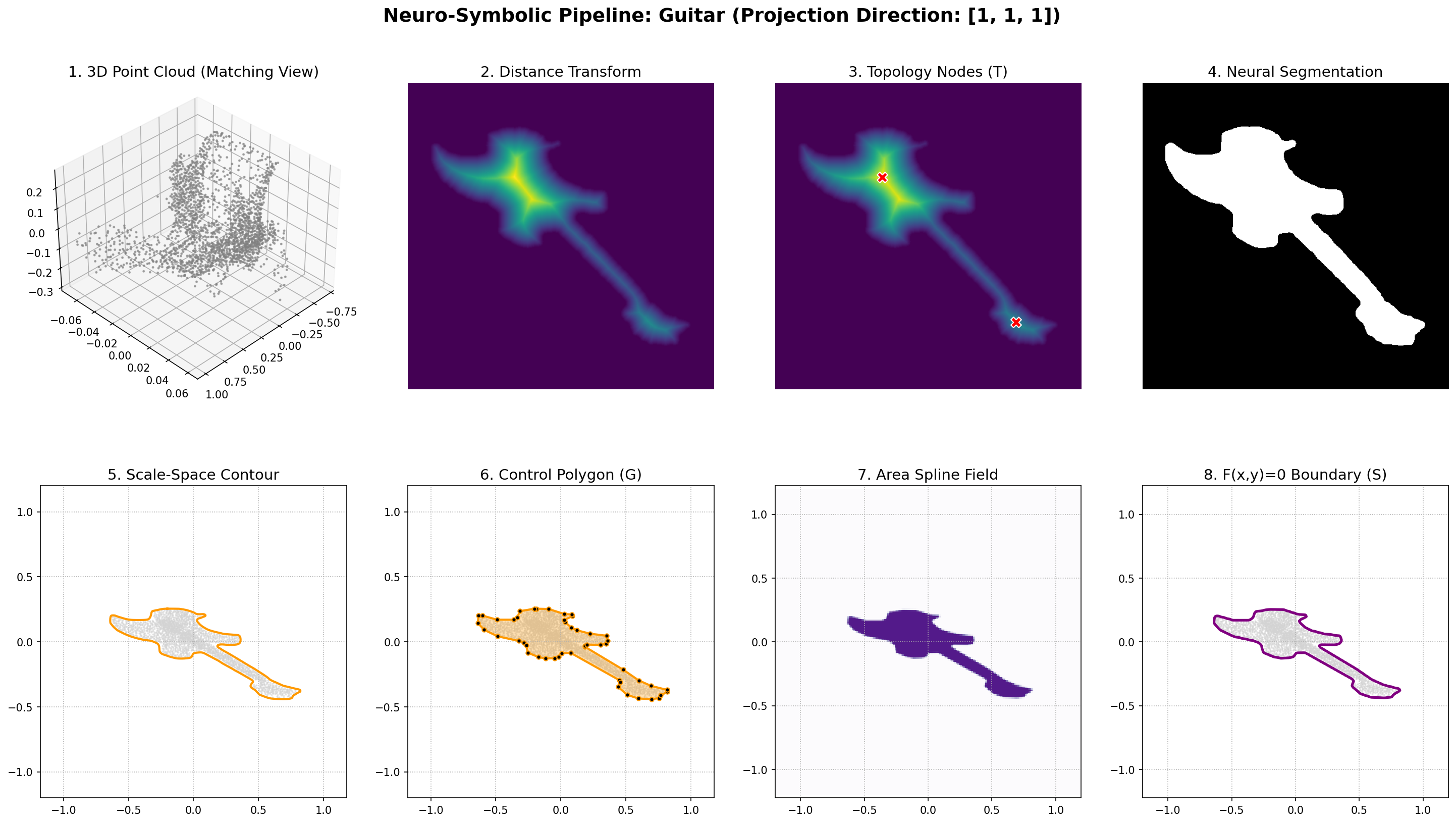}
\includegraphics[scale=0.14]{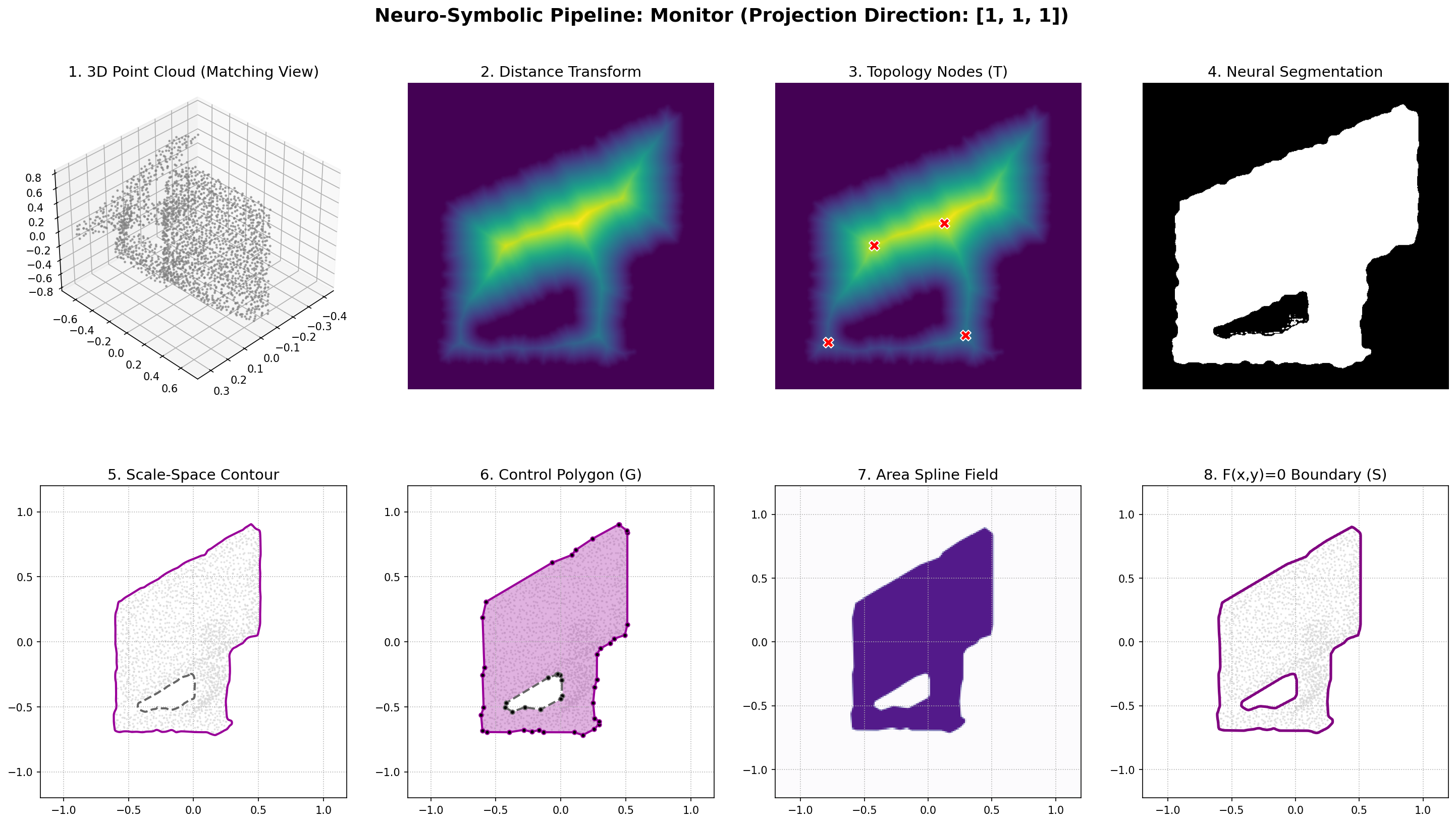} \includegraphics[scale=0.14]{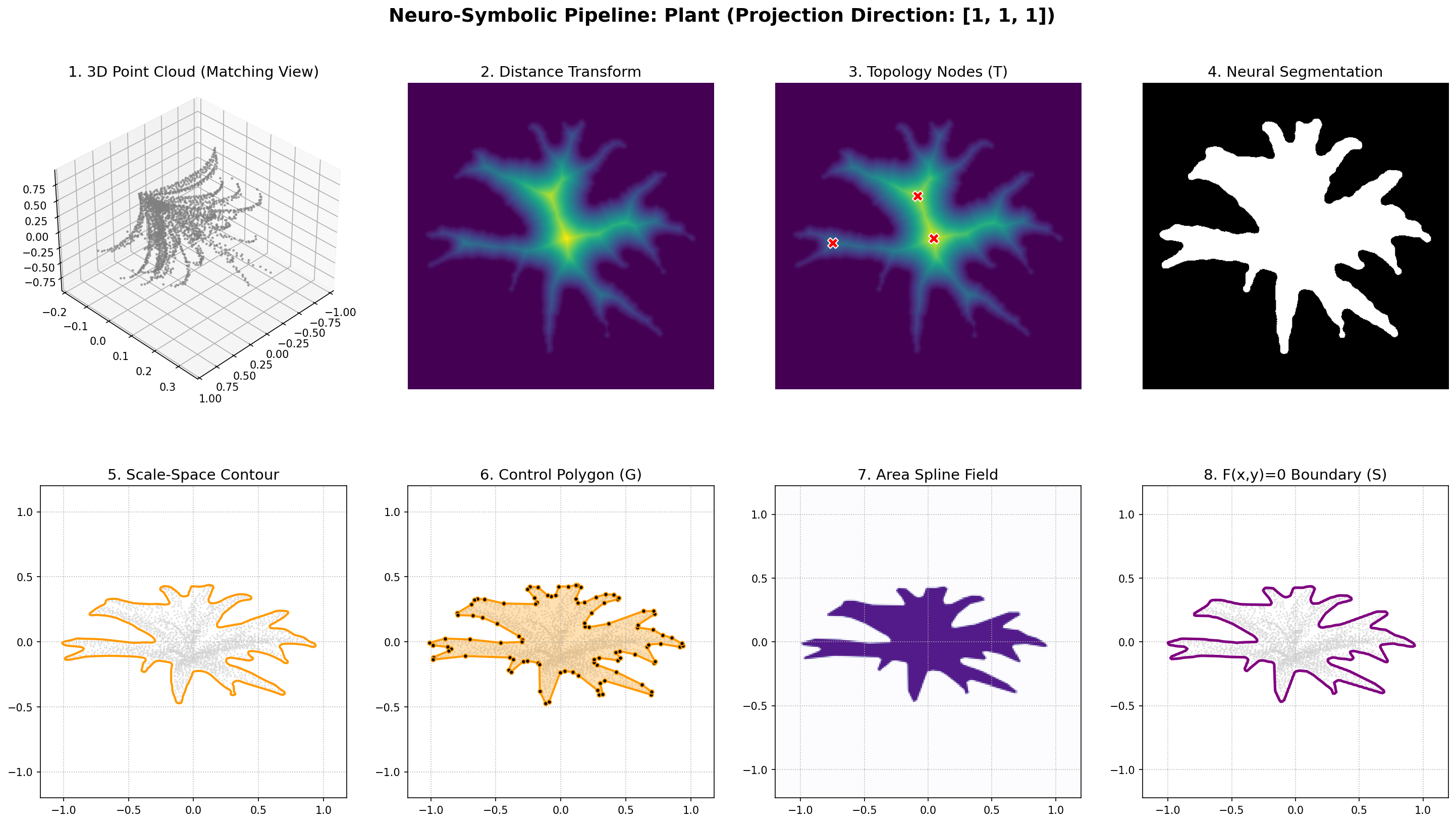}
\includegraphics[scale=0.14]{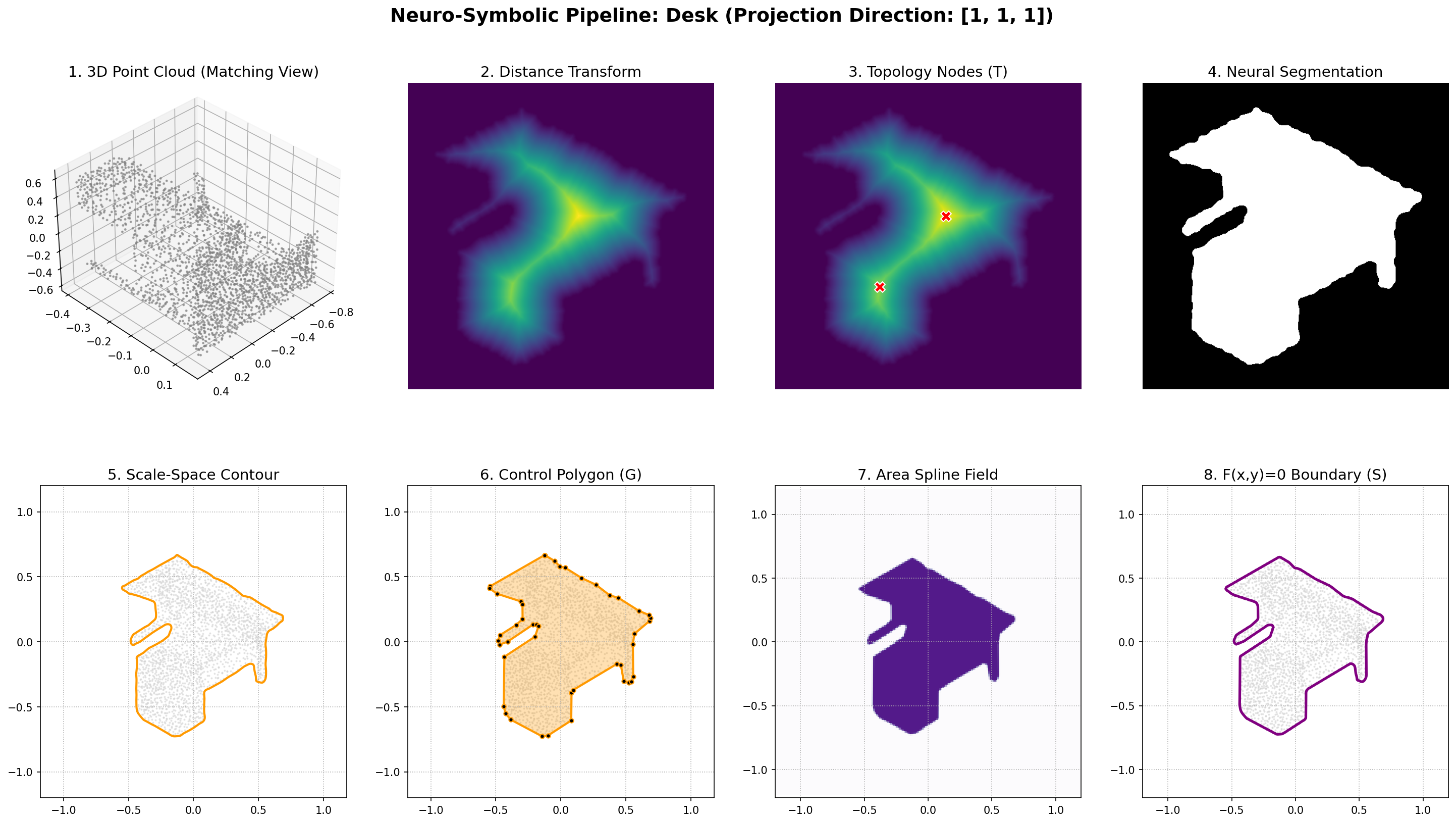} \includegraphics[scale=0.14]{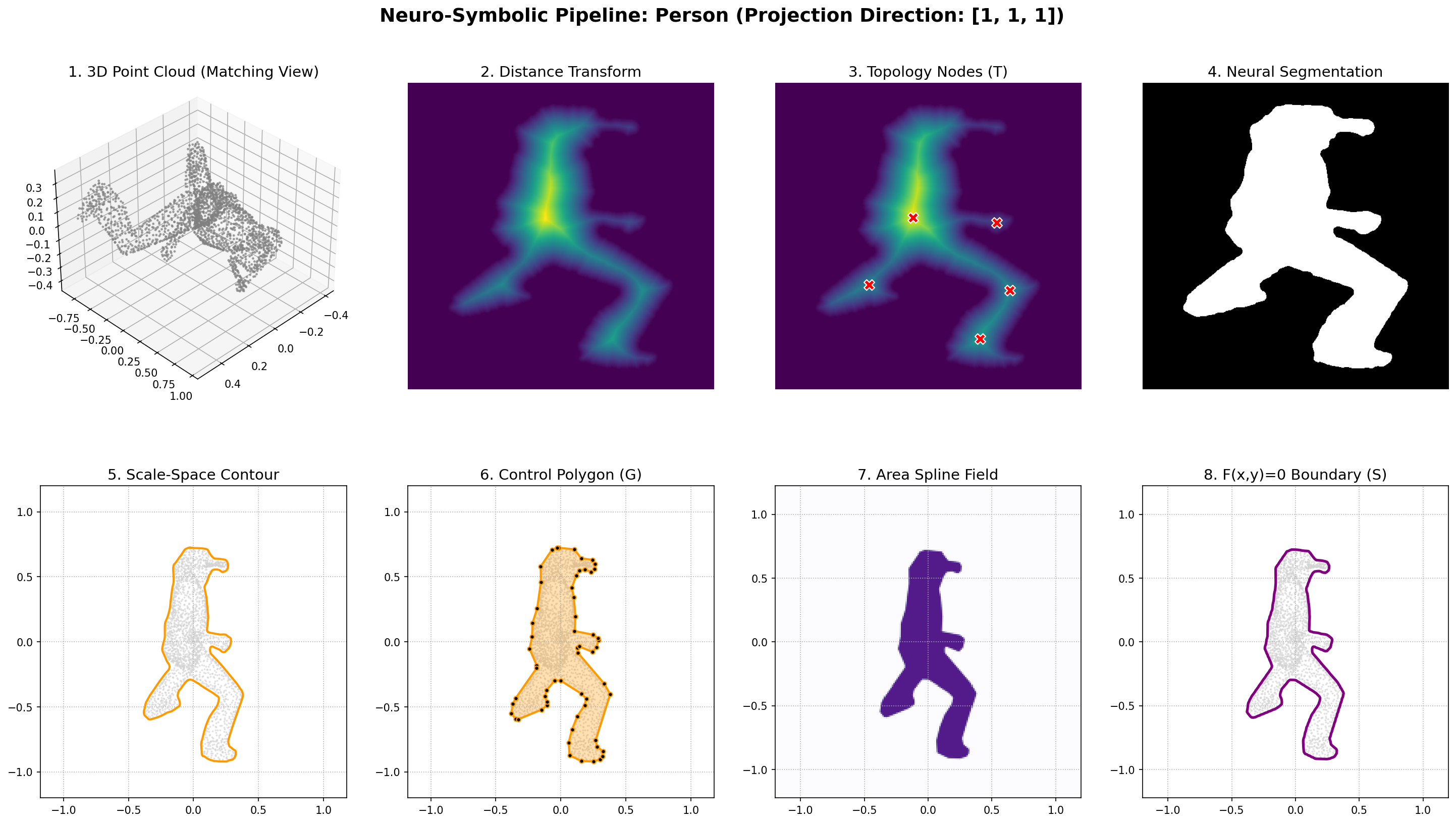}
\includegraphics[scale=0.14]{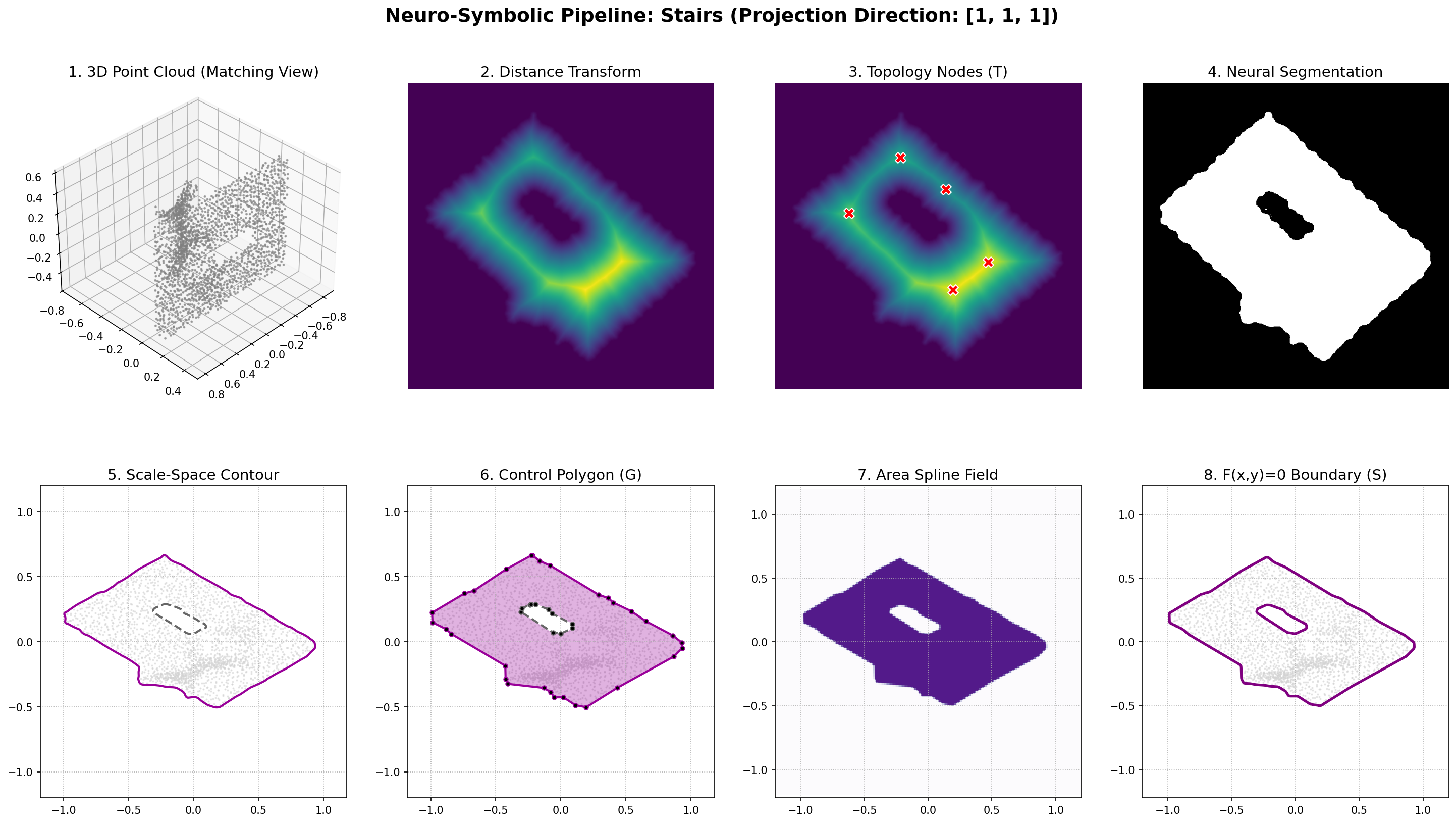} \includegraphics[scale=0.14]{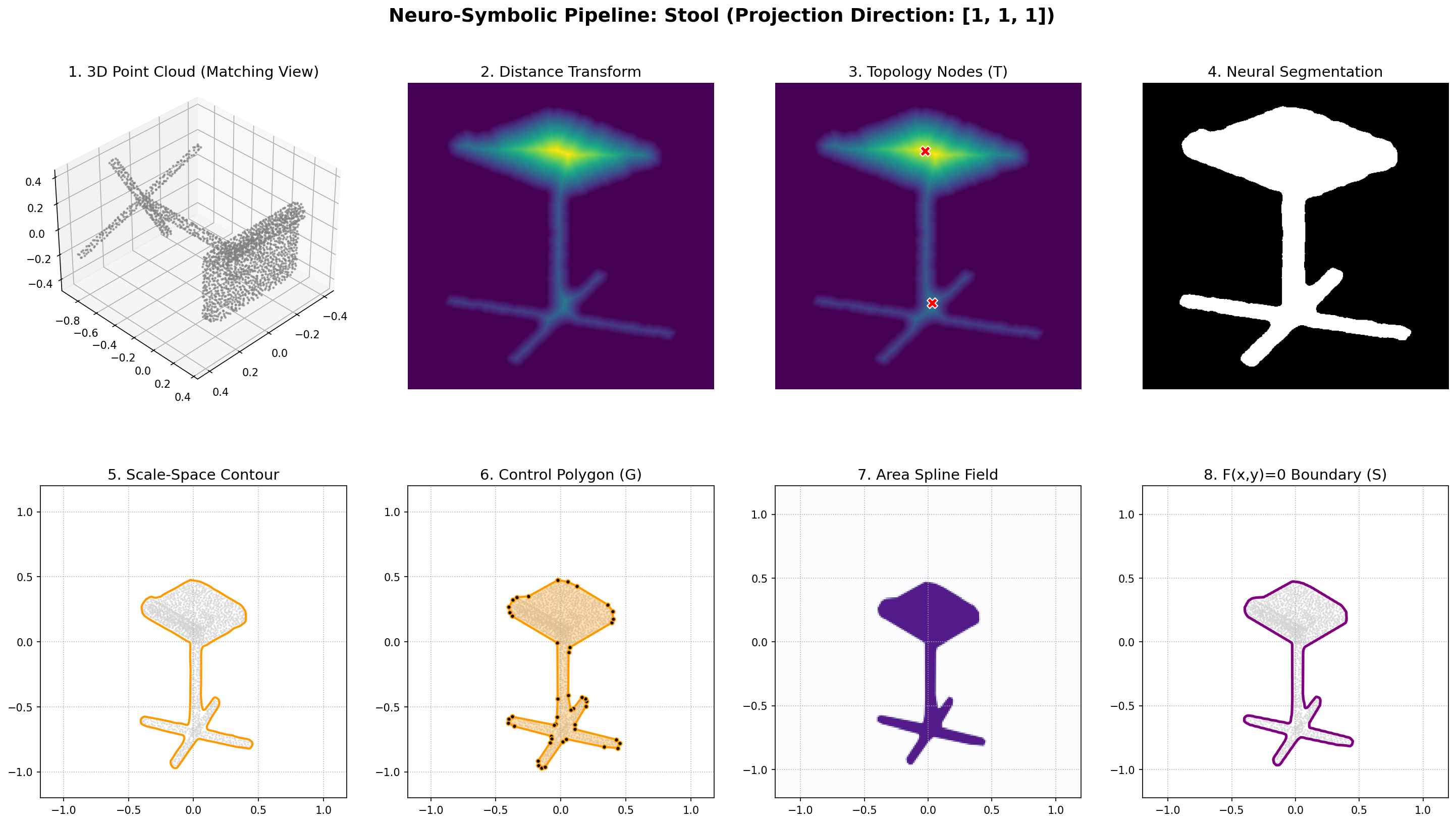}
\caption{ Viewpoint robustness evaluation using arbitrary  projections of ModelNet40 point-cloud objects. The figure illustrates the complete observation-to-symbol transformation ($\mathcal{O}\rightarrow\mathcal{T}\rightarrow\mathcal{G}\rightarrow\mathcal{S}$) for airplane, car, chair, guitar, monitor, plant, desk, person, stairs, and stool models. Each object is processed through the stages of projected observation, Euclidean Distance Transform, topology-aware structural core extraction, topology-guided segmentation, adaptive contour abstraction, sparse geometric control polygon generation, Area Spline field construction, and final symbolic representation $F(x,y)=0$. Unlike canonical top, front, and side projections, the arbitrary viewing direction introduces more complex contour interactions, overlapping structures, and mixed surface visibility. The successful generation of compact symbolic representations demonstrates that the proposed abstraction framework remains stable across substantial viewpoint variations. }
\label{fig:View-111_results} \end{figure*}

The significance of these results extends beyond the recovery of geometric shape. The proposed framework demonstrates a complete transformation from observation to symbol, converting raw sensory measurements into explicit mathematical representations. Viewed from this perspective, the central contribution is not simply reconstruction accuracy, but the formation of symbolic structures from perceptual data. This raises two fundamental questions: why are symbolic representations essential for abstraction, and how can they provide a foundation for explainable machine intelligence? The following sections explore these questions from both cognitive and computational perspectives.

\subsection{Why Symbolic Expressiveness in 2D Matters} At first glance, the use of two-dimensional symbolic representations may appear to be a simplification of the broader three-dimensional reconstruction problem. However, from the perspective of both cognitive science and computer vision, symbolic representation in 2D occupies a fundamentally important position within the abstraction hierarchy. Marr's influential theory of vision proposes that visual understanding proceeds through a sequence of intermediate representations beginning with a two-dimensional primal sketch, followed by a 2.5D sketch, and ultimately a full three-dimensional model \cite{marr1982vision}. Within this framework, geometric understanding does not emerge directly from sensory measurements. Instead, the visual system first extracts structured symbolic descriptions of contours, boundaries, edges, and local geometric organization before constructing higher-level spatial representations. Similarly, Biederman's Recognition-by-Components theory argues that human object recognition relies upon structural information recoverable from two-dimensional contour configurations \cite{biederman1987recognition}. Curvature, symmetry, collinearity, concavity, and boundary relationships provide sufficient information for robust human recognition even when images are degraded, incomplete, or viewed from novel orientations. These findings suggest that contour-level symbolic abstractions are not merely low-level visual features but constitute essential cognitive representations supporting higher-level reasoning. The proposed framework follows a similar philosophy. Rather than attempting direct reconstruction of complete three-dimensional CAD models, the system first transforms observations into explicit two-dimensional symbolic representations through topological abstraction and Area Spline synthesis. The resulting implicit functions \[ F(x,y)=0 \] are mathematically explicit objects possessing semantic structure, analytical properties, and compositional behaviour. Unlike latent neural representations, these symbols can be directly interpreted, manipulated, and reasoned about. From this perspective, the objective of the current experiments is not simply shape reconstruction. Instead, the experiments validate the ability of the proposed architecture to perform the transformation \[ \mathcal{O} \rightarrow \mathcal{T} \rightarrow \mathcal{G} \rightarrow \mathcal{S}, \] where symbolic representations emerge directly from observations. The successful generation of compact Area Spline representations therefore demonstrates the formation of explicit geometric symbols, which may subsequently serve as the foundation for higher-level reasoning, concept formation, and symbolic manipulation.

It is important to note that the symbolic significance of the resulting representations does not arise solely from their ability to reconstruct object boundaries. Conventional parametric spline formulations can also represent smooth geometric contours. However, the objective of the proposed framework is not merely boundary recovery but symbolic representation formation. By representing geometry as an analytical implicit field \[ F(x,y)=0, \] the Area Spline formulation encodes spatial regions, topological relationships, and compositional structure within a single mathematical representation. Combined with its additive field formulation, this allows geometric entities to be composed and manipulated symbolically through algebraic operations, making the representation particularly suitable as the symbolic layer of the proposed neuro-symbolic architecture.

\subsection{Explainability Through Symbolic Representation}
Unlike conventional deep learning systems, whose internal representations are largely encoded within latent model parameters, the proposed architecture maintains explicit representations throughout the abstraction hierarchy. Topological primitives are observable through the EDT field, geometric abstractions are represented by sparse control polygons, and symbolic representations are expressed through closed-form analytical Area Spline formulations. As a result, the complete transformation \[ \mathcal O \rightarrow \mathcal T \rightarrow \mathcal G \rightarrow \mathcal S \] remains fully interpretable. This transparency provides a form of intrinsic explainability rather than post-hoc explanation. The resulting symbolic representations can be inspected, edited, differentiated, composed, and mathematically analysed without reference to hidden neural parameters. Consequently, the proposed framework suggests a pathway toward explainable geometric intelligence through symbolic representation formation.

\subsection{Discussion} The experimental results demonstrate the feasibility of the proposed neuro-symbolic abstraction framework NeuSOGA across multiple object classes and viewing directions. Several observations are particularly important. First, the framework operates without task-specific geometric training. Structural understanding emerges through topological abstraction rather than through statistical learning of object categories. Second, topology-guided prompting consistently produces coherent object segmentations that are subsequently refined through deterministic geometric analysis. Third, adaptive multi-scale abstraction successfully balances global smoothness and local geometric detail. The resulting control polygons remain compact while preserving visually significant structural features. Most importantly, the final output of the pipeline is an explicit symbolic mathematical representation. Unlike neural implicit representations, whose geometry is encoded within network weights, the proposed approach generates analytical Area Spline fields possessing exact mathematical structure, arbitrary-order smoothness, and additive compositional properties. These results therefore support the central hypothesis of this work: that observations can be transformed into symbolic mathematical representations through a neuro-symbolic pipeline that combines topology-driven abstraction, foundation-model perception, geometric simplification, and analytical spline synthesis. 

The experiments also highlight the importance of the chosen symbolic representation. While alternative representations such as Bézier curves, B-splines, or NURBS could be fitted to the extracted contours, these formulations primarily describe boundary parameterizations. In contrast, the proposed Area Spline formulation represents spatial regions directly through an analytical implicit field and supports additive symbolic composition. Consequently, the final representation acts not only as a geometric model but also as an explicit symbolic object suitable for subsequent reasoning and manipulation.

\subsection{Limitations and Future Directions} The experiments presented in this work demonstrate the feasibility of transforming observations into symbolic mathematical representations through the proposed neuro-symbolic abstraction pipeline. However, the current framework focuses primarily on symbolic representation formation and therefore addresses only one stage of a broader cognitive abstraction process. Within the proposed architecture, the Area Spline formulation serves as the symbolic endpoint of the transformation \[ \mathcal{O} \rightarrow \mathcal{T} \rightarrow \mathcal{G} \rightarrow \mathcal{S}. \] The resulting representations are analytical, interpretable, and compositional, but the current system does not yet perform higher-level reasoning directly on these symbols. In particular, automatic discovery of geometric properties, structural invariants, symbolic relations, and conceptual hierarchies remains an open problem. Future work will therefore investigate mechanisms operating on top of the symbolic representation layer. Potential directions include automatic symmetry discovery, repeated motif identification, structural decomposition, shape grammar formation, and symbolic concept learning from collections of geometric observations. Such capabilities would extend the present framework from symbolic representation formation toward the broader objectives of concept formation and machine-assisted mathematical reasoning. A second limitation concerns experimental scale. Although the current results demonstrate robustness across multiple object categories, viewing directions, and sensing modalities, future evaluation should investigate larger and more diverse datasets. Particular attention will be given to assessing the stability of symbolic representations under changes in observation quality, sensing modality, occlusion, and geometric complexity. More broadly, the present work establishes a symbolic representation layer for geometric intelligence. Future research will explore how such representations may serve as the foundation for explainable reasoning systems capable of constructing increasingly abstract geometric knowledge from observations.

\section{Conclusion} This paper introduced NeuSOGA, a neuro-symbolic framework for transforming observations into explicit symbolic mathematical representations. Inspired by theories of human abstraction and symbolic reasoning, the proposed architecture implements a hierarchy of representation formation \[ \mathcal{O} \rightarrow \mathcal{T} \rightarrow \mathcal{G} \rightarrow \mathcal{S}, \] where observations are progressively transformed into topological abstractions, geometric abstractions, and ultimately symbolic mathematical representations. Unlike conventional learning-based reconstruction systems that encode geometric knowledge within latent neural parameters, the proposed framework separates perception, abstraction, and symbolic representation into distinct computational stages. Topology-aware structural primitives are first extracted through Euclidean Distance Transforms and subsequently used to guide foundation-model perception. Adaptive multi-scale geometric abstraction then converts dense observations into sparse structural representations, which are finally projected into analytical symbolic representations through the Implicit Area Spline formulation. The experimental results demonstrated that this abstraction process remains effective across multiple object categories, sensing modalities, and viewing directions. Evaluations on ModelNet40 point-cloud data, arbitrary-view projections, and segmented optical observations confirmed that the same computational architecture can consistently generate compact symbolic representations while preserving essential geometric and topological structure. These findings suggest that the proposed framework operates on abstract spatial organization rather than modality-specific characteristics. A central contribution of this work is the identification of the Implicit Area Spline as a symbolic representation layer. Unlike conventional parametric boundary representations, the Area Spline formulation represents spatial regions directly through analytical implicit fields, supports arbitrary-order smoothness, and admits additive symbolic composition. The resulting representation is not merely a reconstructed shape but an explicit mathematical object that can be inspected, manipulated, and analysed without reference to hidden neural parameters. From a broader perspective, the proposed framework contributes to Neuro-Symbolic AI by demonstrating a pathway through which symbolic representations can emerge from perceptual observations. Because each stage of the abstraction hierarchy remains observable and interpretable, the architecture also provides a form of explainability by construction, in contrast to the largely latent representations employed by modern deep learning systems. The present work focuses on symbolic representation formation as a fundamental prerequisite for higher levels of machine intelligence. Future research will investigate mechanisms capable of operating directly on the generated symbolic representations to discover geometric properties, structural invariants, compositional relationships, and higher-level concepts. From this perspective, the proposed framework establishes a symbolic foundation for future systems that move beyond perception toward explainable reasoning, concept formation, and machine-assisted mathematical discovery.

\bibliographystyle{plain}
\bibliography{references}

\end{document}